\documentclass[10pt,twocolumn,letterpaper]{article}

\usepackage{cvpr}              

\usepackage{CJKutf8}

\usepackage[table]{xcolor}
\usepackage{booktabs,multirow,array}
\usepackage{makecell}
\usepackage{tikz}
\usepackage{arydshln} 
\usepackage{colortbl}
\usepackage{amsthm}
\usepackage{amsmath, amssymb}
\usepackage{adjustbox}
\usepackage{changepage}
\usepackage{tabularx}
\usepackage{lipsum}
\usepackage{paralist}
\usepackage{comment}
\usepackage{graphicx}
\usepackage{subcaption} 
\usepackage{pifont}
\usepackage{fontawesome}
\usepackage{xspace}
\usepackage{float}
\usepackage{algorithm}
\usepackage{listings}
\usepackage{algcompatible}
\usepackage{algpseudocode}
\usepackage{soul}
\usepackage{rotating}
\usepackage{enumitem}
\usepackage{bm}
\usepackage[most]{tcolorbox}
\usepackage{marvosym}

\definecolor{lightblue}{RGB}{216,224,255}
\definecolor{lightgrey}{RGB}{240,240,240}
\newcommand{\customfont}{\linespread{0.9}\selectfont}

\definecolor{cvprblue}{rgb}{0.21,0.49,0.74}
\usepackage[pagebackref,breaklinks,colorlinks,allcolors=cvprblue]{hyperref}

\title{Modeling Cross-vision Synergy for Unified Large Vision Model}

\author{Shengqiong Wu$^1$\thanks{Email: swu@u.nus.edu}\;, Lanhu Wu$^1$, Mingyang Bao$^1$, Wenhao Xu$^1$, \\ 
Hanwang Zhang$^2$, Shuicheng Yan$^1$, Hao Fei$^{1,\textrm{\Letter}}$\;, Tat-Seng Chua$^1$\\
$^1$National University of Singapore, $^2$Nanyang Technological University, 
}

\begin{document}
\maketitle

\def\thefootnote{\textrm{\Letter}}\footnotetext{Corresponding Author.}

\begin{abstract}
Recent advances in large vision models (LVMs) have shifted from modality-specific designs toward unified architectures that jointly process images, videos, and 3D data. 
However, existing unified LVMs primarily pursue functional integration, while overlooking the deeper goal of cross-vision synergy: the ability to reason over complementary priors across visual modalities. 
To address this, we present \textbf{PolyV}, a unified LVM that achieves cross-vision synergy at both the architectural and training levels.
Architecturally, PolyV adopts a sparse Mixture-of-Experts LVM coordinated by a dynamic modality router, allowing each expert to specialize in modality-specific priors while enabling bidirectional interaction and mutual refinement across modalities.
Training-wise, a synergy-aware paradigm combines modality-specific pretraining with coarse-to-fine synergy tuning via knowledge distillation and object-/relation-level alignment.
Extensive experiments on 10 benchmarks spanning image, video, and 3D understanding, including synergy-focused datasets requiring spatial or temporal priors, demonstrate that PolyV consistently outperforms existing models, achieving over 10\% average improvement over its backbone. 
Overall, PolyV establishes a unified framework for synesthetic visual reasoning, advancing toward truly synergistic LVMs. Project page: \href{https://sqwu.top/PolyV}{PolyV.io.}

\end{abstract}

\makeatletter
\let\origaddcontentsline\addcontentsline
\renewcommand{\addcontentsline}[3]{}
\makeatother

\section{Introduction}
\label{sec:intro}

\begin{figure}[t]
  \centering
   \includegraphics[width=0.99
   \linewidth]{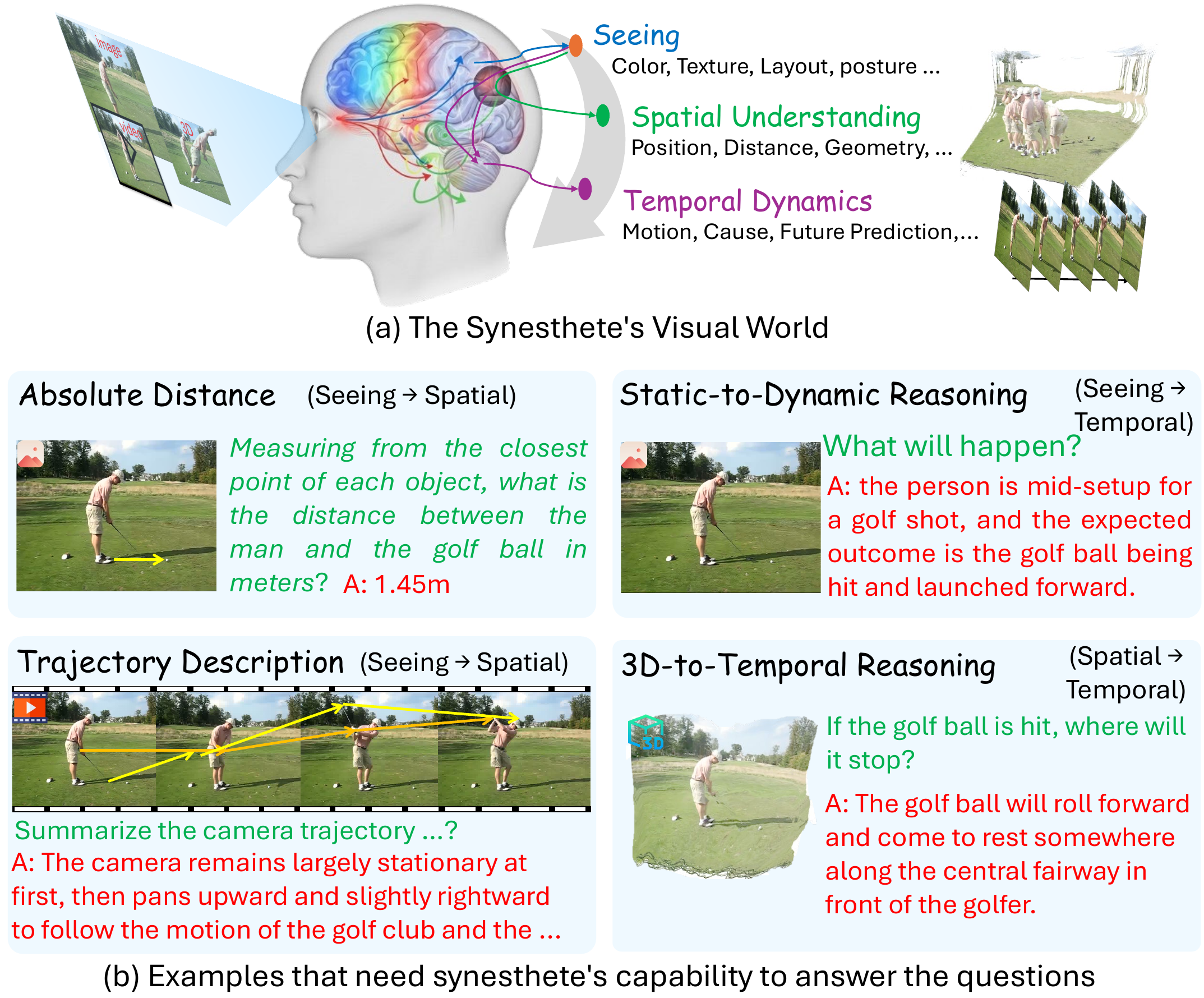}
   \vspace{-2mm}
   \caption{(a) Human perception integrates visual, spatial, and temporal cues synergistically, enabling reasoning across modalities. (b) Examples illustrate such synergy, inferring motion from static images and transferring 3D priors to improve video understanding.}
   \label{fig:intro}
   \vspace{-4mm}
\end{figure}

Large vision models (LVMs)~\cite{liu2024improved,bai2025qwen25vl,hong20233dllm,chen2024ll3da} have demonstrated impressive progress, with a growing trend toward vision unification, building a single model that integrates multiple modalities through unified modeling~\cite{shen2023hugginggpt,han2023imagebindllm}. 
Extensive research attention has been gathered, leading to numerous efforts that integrate vision data in different modalities within a shared framework~\cite{zhang2025uni3dmoe,huang2023embodied}. 
Researchers have gradually realized, however, that the goal of unification extends beyond functional integration: the essential aim is to enable a synergetic learning across modalities, as images, videos, and 3D scenes are inherently different hierarchical forms of the same visual signal and should thus share fundamental visual features. 
Recent works~\cite{zheng2025video3dllm,zhu2025llava3d,wang2025ross3d} explore this idea by transferring video knowledge to 3D scene understanding. 
Yet, as illustrated in Fig. \ref{fig:intro}, existing unified LVMs still fall short of achieving \textbf{true cross-vision synergy}, that is, \textit{synesthetic reasoning across visual modalities}, analogous to the human synesthetic visual system~\cite{barnett2008differences,palmeri2002perceptual,simner2006synaesthesia}.
For instance, temporal and motion priors from videos could inform dynamic inference in static images, while 3D geometric priors could enhance spatial reasoning in videos.
Unfortunately, current research remains biased toward transferring information from lower- to higher-order modalities~\cite{zhu2025llava3d,zheng2025video3dllm,wang2025ross3d}, neglecting the bidirectional and interactive nature of such synergy. 
Consequently, two critical challenges remain underexplored: \textit{(i) the lack of synergy-oriented architectural design} and \textit{(ii) the absence of training strategies that foster cross-vision interaction}.

From an architectural perspective, current unified LVMs lack designs that explicitly facilitate cross-vision synergy. 
Early studies primarily pursued straightforward modality unification, where each modality employed a separate encoder, as seen in~\cite{wu2024nextgpt,zhan2024anygpt,xu2025qwen25omni,abouelenin2025phi,huang2023embodied}.
More recent efforts~\cite{bai2025qwen25vl,zhang2024videollava} have begun to explore shared encoders to process different visual modalities by adopting a single image encoder backbone for both image and video inputs, while other works~\cite{wang2025ross3d,zheng2025video3dllm} extend video-encoder architectures to support 3D representations. 
However, these frameworks still lack mechanisms for genuine inter-modal synergtic modeling: they simply rely on shared encoders or feature concatenation rather than explicit cross-modality interaction modules. 
Consequently, features from different modalities remain isolated, preventing the dynamic exchange of priors during inference. 
A truly synergistic architecture should instead disentangle and coordinate modality-shared and modality-specific features, allowing information to flow freely across modalities and establishing the foundation for authentic cross-vision reasoning.

From the training perspective, current LVMs lack synergy-aware learning.
Most existing approaches rely solely on supervised fine-tuning~\cite{lin2024videollava,zhang2024videollava}: either training each modality-specific module independently, thereby preventing any feature sharing, or incrementally fine-tuning across modalities~\cite{wang2025ross3d,zheng2025video3dllm}, which largely results in catastrophic forgetting and weak cross-modal generalization. 
Achieving true synergy-aware training requires joint multimodal learning, where the model simultaneously observes and aligns data from multiple visual modalities to maintain a coherent, shared-specific dual representation space rather than only fragmented, modality-specific ones.
Furthermore, enabling deeper synergy demands moving beyond coarse, instance-level alignment toward fine-grained correspondence modeling, i.e., object-level attributes, spatial relations, and event consistency.
Incorporating such fine-grained alignment mechanisms can substantially strengthen cross-modal reasoning, allowing the model to generalize ``by analogy'' across visual modalities.

To bridge the aforementioned gaps, we propose a unified framework, named \textbf{PolyV}, for building a cross-vision synergistic LVM. 
As illustrated in Fig.~\ref{fig:framework}, our design adopts a sparse Mixture-of-Experts (MoE) architecture, where a dynamic router coordinates multiple experts to enable each expert to refine its domain knowledge while preserving complementary priors learned by others, thereby preventing knowledge degradation and fostering bidirectional synergy across modalities.
On top of this architecture, we further introduce a synergy-aware training strategy comprising two stages:
\textbf{(1) Modality-specific pretraining}, where each expert independently learns modality-specific representations to establish strong foundational knowledge; and
\textbf{(2) Synergy-aware fine-tuning}, which includes coarse-grained synergy tuning (i.e., distilling instance-level knowledge from external teacher models to the target experts), and fine-grained synergy tuning, aligning object- and relation-level representations across modalities to enable precise and holistic cross-modal synergetic reasoning.
After training, PolyV can take input from a single visual modality while implicitly leveraging modality-specific priors learned by all experts through synesthetic reasoning, thus extending beyond the information of the observed modality and achieving robust cross-vision understanding.

We conduct extensive experiments across ten representative benchmarks covering diverse visual modalities, including standard image, video, and 3D understanding tasks, as well as synergy-oriented benchmarks that require strong spatial or temporal priors.
Experimental results show that our proposed PolyV consistently outperforms prior models across all settings, achieving an average improvement of over 10\% compared with its backbone VLM.
Comprehensive ablation studies further validate the effectiveness of our MoE architecture and synergy-aware training strategy, demonstrating their essential roles in enhancing cross-vision reasoning.
Moreover, analysis of expert routing distributions reveals that PolyV adaptively leverages modality-specific priors and dynamically allocates experts across tasks, achieving efficient and interpretable path routing.
Overall, this work presents the first unified large vision model explicitly designed for cross-vision synergy, providing a general and extensible framework for structured multimodal reasoning across image, video, and 3D domains.

\vspace{-1mm}
\section{Related Work}

\vspace{-2mm}
\paragraph{Large Vision Model for Vision Unification.}

Recent advances in LVMs~\cite{liu2024improved,zhang2023nextchat} have sparked significant progress by integrating vision modalities into a large language model (LLM) through a projection layer.
Early studies focused on single-modality modeling, such as image~\cite{liu2024improved}, video~\cite{lin2024videollava,zhang2024videollava}, and 3D~\cite{hong20233dllm,chen2024ll3da,xu2024pointllm}. 
Subsequently, researchers began incorporating multiple modalities into a unified framework by employing separate encoders for each modality and combining their representations through straightforward feature concatenation~\cite{wu2024nextgpt,xu2025qwen25omni}. 
While this approach achieved functional unification, allowing a single system to handle heterogeneous inputs, it failed to capture the deeper interdependence among visual modalities.
Later studies recognized that these modalities are not only complementary but also intrinsically correlated, prompting attempts to leverage cross-modal priors, such as adapting image-based models for video understanding or transferring video-based knowledge to 3D scene modeling~\cite{wang2025ross3d,zheng2025video3dllm,zhu2025llava3d}.
Nevertheless, these approaches remain limited: architecturally, existing models lack mechanisms to facilitate effective feature sharing and interaction across modalities; moreover, simple supervised fine-tuning or incremental training fails to achieve robust knowledge preservation and alignment among modalities.

\paragraph{Mixture-of-Expert Structure.}

The MoE architecture consists of multiple expert networks, each responsible for modeling a distinct subspace of knowledge.
By adaptively activating only a subset of experts during inference, MoE architectures significantly reduce computational overhead while improving representational capacity and reasoning efficiency~\cite{huang2024toward}.
MoE has been successfully applied across a variety of domains. 
In natural language processing, MoE has been applied for multilingual language models~\cite{lepikhin2020gshard}, and large-scale multi-task settings~\cite{du2022glam}.
In the visual domain, MoE-based structures have been explored in unimodal understanding, including image understanding~\cite{lin2024moellava,shen2024mome,xin2025i2moe,zhu2024unimed}, video perception~\cite{nguyen2025temporal,yang2025astrea}, and 3D scene understanding~\cite{zhang2025uni3dmoe}.
Later, multimodal MoE~\cite{li2025unimoe} further validates the effectiveness of expert modularization in enhancing generalization and cross-modal reasoning.
Collectively, prior research underscores MoE's versatility in scalability and efficacy.
In this work, we extend the MoE to develop a unified LVM that enables cross-vision synergy across image, video, and 3D modalities.

\section{Method}

\subsection{Overview}
\begin{figure}[!t]
  \centering
   \includegraphics[width=0.99\linewidth]{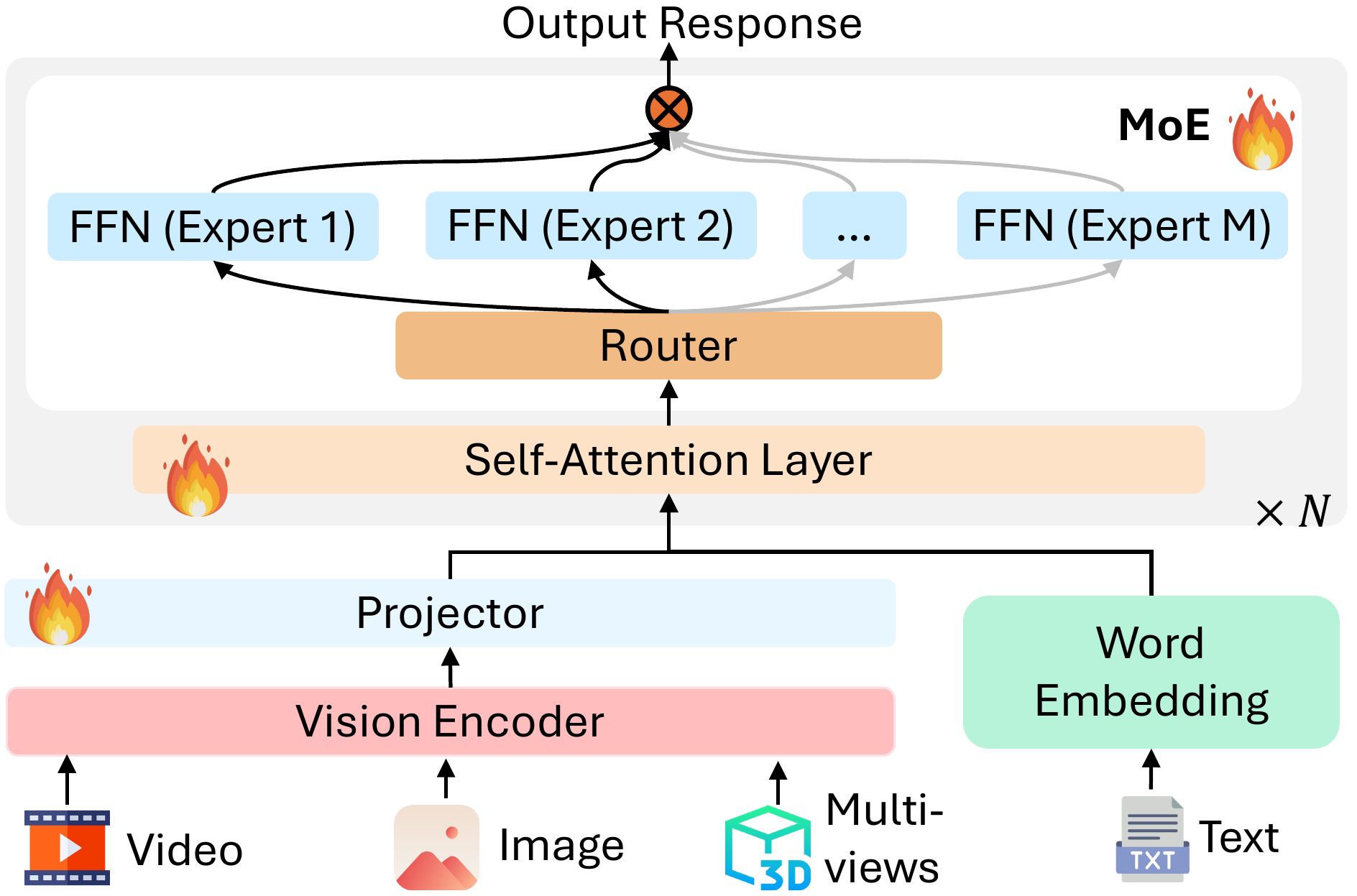}
   \vspace{-2mm}
   \caption{An illustration of \textbf{PolyV}, where an MoE architecture is designed to enable synergistic learning across image, video, and 3D modalities. Fire denotes the trainable parameters.}
   \label{fig:framework}
   \vspace{-4mm}
\end{figure}

Motivated by the goal of achieving cross-vision synergistic learning, that is, fully exploiting the complementary and distinctive information across visual modalities, we design \textbf{PolyV}, as illustrated in Fig.~\ref{fig:framework}.
The model comprises a universal vision encoder,  a word embedding layer, a projection layer, and multiple stacked LLM blocks integrated with MoE blocks. 
Following, we describe the model architecture in Section~\ref{sec:arhitecture} and the training strategy in Section~\ref{sec:training}.

\subsection{Architecture of PolyV}
\label{sec:arhitecture}

\paragraph{Universal Vision Encoder.}

Given an arbitrary visual input $\mathcal{I}=\{I_1, \cdots, I_K\}$, where an image corresponds to a single frame, a video represents a sequence of frames, and 3D data consists of multiple rendered views, a universal vision encoder~\cite{dosovitskiy2020image} is employed to extract a visual token sequence $\bm{V} \in \mathbb{R}^{p \times c}$.
While sharing a common architecture, the encoder is adapted slightly for different visual modalities.
For video inputs, we treat the video as a temporally ordered sequence of frames.
A dynamic frame sampling strategy is applied to select representative key frames, and temporal positional embeddings are added to capture temporal dynamics and maintain motion coherence.
Following~\cite{zhu2025llava3d,wang2025ross3d},  for 3D inputs, the additional per-pixel 3D coordinates using a positional encoding function are encoded and injected into the visual features to enhance geometric awareness.
The resulting visual representations are then projected through a visual projector that maps $\bm{V} \in \mathbb{R}^{P \times C}$ to $\bm{H} \in \mathbb{R}^{P \times D}$, where $D$ denotes the hidden dimensionality of the LLM.
Similarly, textual inputs are embedded through a word embedding layer, yielding textual tokens $\bm{T} \in \mathbb{R}^{N \times D}$, where $N$ is the text sequence length.

\begin{figure*}[!t]
\centering
\includegraphics[width=0.99\textwidth]{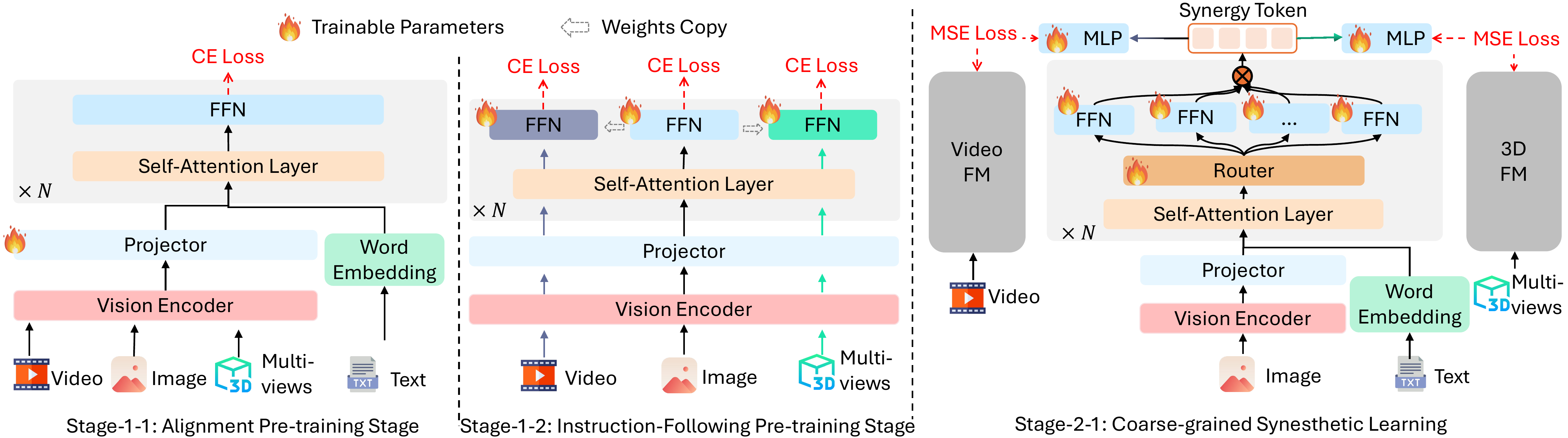}
\vspace{-2mm}
\caption{
Illustration of detailed training stages. 
Stage-1(-1/2) focuses on enabling model understanding of each vision modality. 
Stage-2(-1): introduces coarse-grained synergistic learning, where a video and 3D foundation model distill temporal and geometric priors into the MoE-LLM.
During this process, the model generates latent synergy tokens wrapped in \texttt{<synergy>}, which are optimized via MSE loss to align with the knowledge extracted from foundation models, thereby fostering cross-modality reasoning.
}
\label{fig:training}
\vspace{-3mm}
\end{figure*}

\vspace{-2mm}
\paragraph{MoE Structure.}
We concatenate the visual token sequence and text token sequence, and feed the combined representation into an LLM composed of stacked multi-head self-attention (MSA) and feed-forward network (FFN) layers.
To implement the MoE architecture, each standard FFN layer is expanded into multiple parallel expert networks, forming an ensemble of experts $\mathcal{E} = \{E_1, \cdots, E_M\}$, where $M$ is the number of experts.
The computation across $L$ layers can be formulated as:
{
\setlength{\abovedisplayskip}{3pt}
\setlength{\belowdisplayskip}{3pt}
\begin{align}
\bm{X}_0 &= [\bm{H}, \bm{T}], \label{eq:input} \\
\bm{X}'_{\ell} &= \mathrm{MSA}(\mathrm{LN}(\bm{X}_{\ell-1})) + \bm{X}_{\ell-1}, \quad \ell = 1, \ldots, L, \label{eq:msa} \\
\bm{X}_{\ell} &= \mathrm{MoE}(\mathrm{LN}(\bm{X}'_{\ell})) + \bm{X}'_{\ell}, \quad \ell = 1, \ldots, L, \label{eq:moe} \\
\bm{Y} &= \mathrm{LN}(\bm{X}_L). \label{eq:output}
\end{align}
}
In each MoE layer, a sparse router determines which experts should process each token. Specifically, the router is a linear projection that predicts the probability of assigning a token to each expert. For computational efficiency, only the top-$k$ experts with the highest routing probabilities are activated for each token, and their outputs are combined using a weighted sum based on the normalized routing scores:
{
\setlength{\abovedisplayskip}{3pt}
\setlength{\belowdisplayskip}{3pt}
\begin{align}
    \mathcal{P}(\bm{X}_{\ell})_{i} &= 
\frac{e^{f(\bm{X}_{\ell})_{i}}}
{\sum_{j}^{M} e^{f(\bm{X}_{\ell}^{s})_{j}}}, \\
\mathrm{MoE}(\bm{X}_{\ell}) &= 
\sum_{i=1}^{k} \mathcal{P}(\bm{X}_{\ell})_{i} \cdot \mathcal{E}(\bm{X}_{\ell})_{i}.
\end{align}
}

\subsection{Training Process}
\label{sec:training}

\paragraph{Stage-1: Modality-specific pretraining.}
This stage aims to enable the model to understand each specific visual modality.
\textbf{First}, we establish an initial vision-language connection by optimizing the projector, which maps multi-modal inputs into soft tokens within the language space.
The model is trained on captioning tasks using the cross-entropy loss $\mathcal{L}{ce}$, as shown in Fig.~\ref{fig:training}(a).
After acquiring a preliminary understanding of visual content, each expert specializes in modality-specific properties, such as temporal dynamics in videos and spatial geometry in 3D scenes, using more complex vision–language instruction data (Fig.~\ref{fig:training}(b)).
While $\mathcal{L}{ce}$ remains the primary objective, the FFN layers are further adapted to align with the intrinsic characteristics of each modality.

\vspace{-3mm}
\paragraph{Stage-2: Synergy-aware Training.}
This stage aims to endow the model with cross-vision synergy, i.e., the ability to perform synesthetic reasoning across visual modalities.
The modality-specific FFNs from Stage 1 are integrated into MoE layers, where a dynamic router enables mutual complementation among experts.
Beyond this implicit architectural synergy, we introduce explicit \texttt{<synergy>} tokens that act as cognitive mediators, encouraging the model to form an intermediate ``mental-state'' representation before producing the final response.
We conceptualize visual synergy at two complementary levels:
\textbf{(i)} coarse-grained, instance-level understanding of global scene context; and
\textbf{(ii)} fine-grained alignment of object- and relation-level correspondences across modalities.
Accordingly, our synergy-aware training consists of two components: coarse-grained synergy tuning and fine-grained synergy tuning.

\begin{figure*}[!t]
  \centering
   \includegraphics[width=0.99\textwidth]{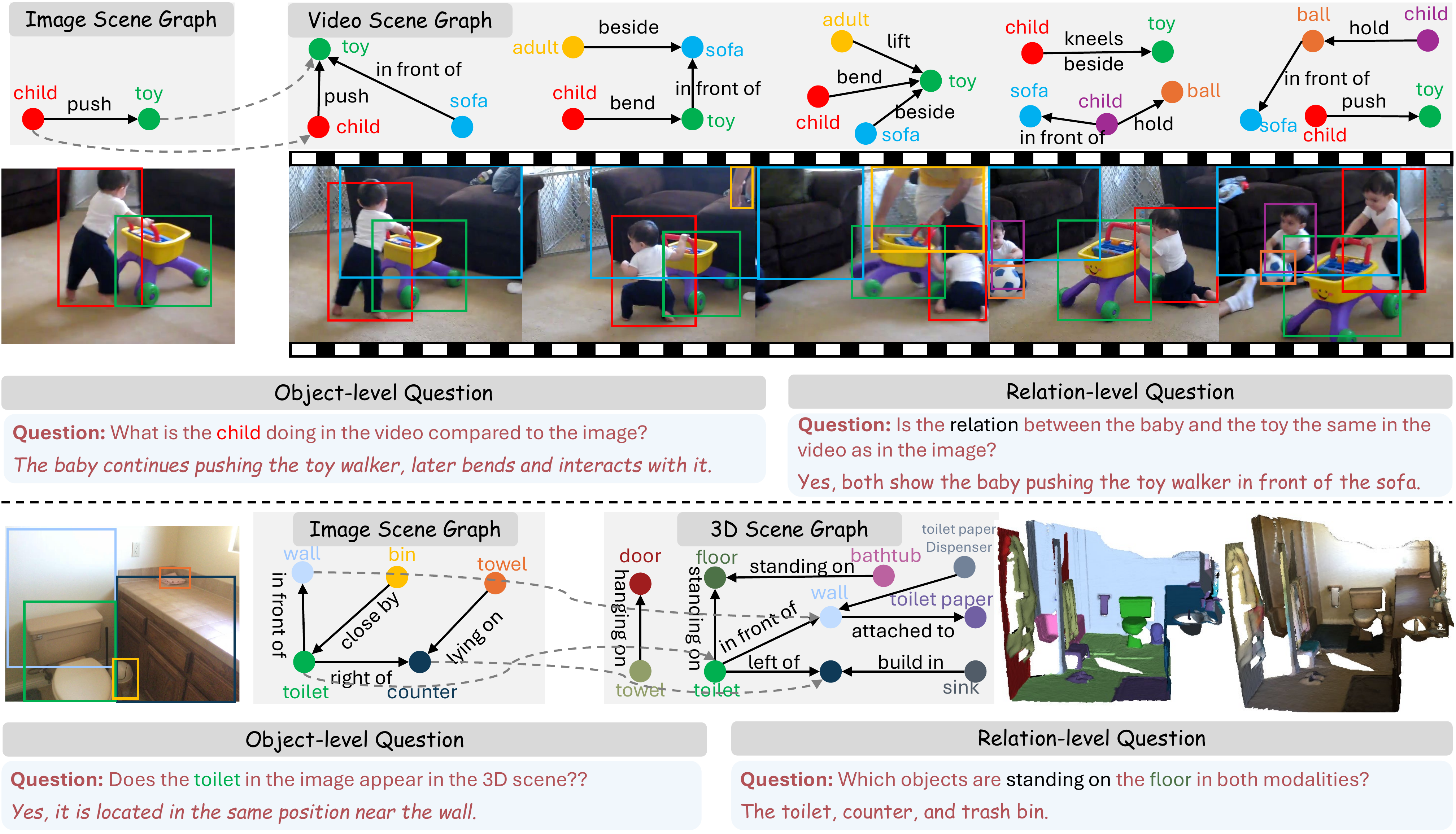}
   \vspace{-2mm}
   \caption{
   Illustration of cross-vision synergy question-answer pairs. Inspired by~\citep{wu2025usg}, we leverage the universal scene graph constructed from image-video and image-3D (multi-view) to construct the object-/relation-level cross-synergy question-answer pairs, which are then utilized to enable the model to achieve fine-grained cross-vision synergy.
   }
   \label{fig:fine-grained}
   \vspace{-3mm}
\end{figure*}

\textbf{$\blacktriangleright$ Coarse-Grained Synergy Tuning}
Relying solely on text supervision restricts the model to abstract semantic alignment and prevents learning rich modality-specific priors.
Inspired by~\cite{zhu2025llava3d,chen2025thinkwith3d}, we adopt a knowledge-distillation framework that transfers temporal and spatial knowledge from strong single-modality foundation models into our unified model.
Specifically, a video foundation model (e.g., V-JEPA 2~\cite{assran2025vjepa2}) provides low-dimensional temporal priors, while a 3D foundation model (e.g., VGGT~\cite{wang2025vggt}) encodes geometric and spatial structures.
As illustrated in Fig.~\ref{fig:training}(b), we design two distillation pathways, i.e., video$\rightarrow$image and 3D$\rightarrow$image, to guide the learning.
Given an input image with a corresponding video or 3D textual description, we extract the hidden representations of the \texttt{<synergy>} tokens from the last layer of the MoE-LLM, denoted as $\bm{F}_{\text{syn}}$.
From the teacher models, we obtain the temporal feature $\bm{F}_{\text{temporal}} = f_{\text{VFM}}(V)$ from the final aggregator of the video foundation model, and the spatial feature $\bm{F}_{\text{temporal}} = f_{\text{3DFM}}(3D)$ from the 3D foundation model.
To ensure dimensional consistency, two projectors are employed to map the synergy latent into aligned feature spaces:
\begin{equation}
    \bm{F}^{v} = f_{\text{mlp}}^t(\bm{F}_{\text{syn}}), \;\; \bm{F}^{g} = f_{\text{mlp}}^g(\bm{F}_{\text{syn}})
\end{equation}
The training objective minimizes the discrepancy between the projected synergy features and the corresponding teacher features:
\begin{equation}
\begin{aligned}
     \mathcal{L}_{\text{coarse}} &= \mathcal{L}_{\text{temporal}} +  \mathcal{L}_{\text{spatial}} \\
                &= ||\bm{F}_{\text{temporal}} - \bm{F}^{v}||^2 + ||\bm{F}_{spatial} - \bm{F}^{g}||^2
\end{aligned}
\end{equation}
Additionally, we incorporate a differentiable load-balancing loss~\cite{fedus2022switch} into each MoE layer to encourage balanced token distribution among experts:
\begin{equation}
\mathcal{L}_{\text{aux}} = M \cdot \sum_{i=1}^{M} \mathcal{F}_i \cdot \mathcal{G}_i,
\label{eq:aux_loss}
\end{equation}
where $\mathcal{F}$ represents the fraction of tokens processed by each expert ${E}_i$, and $\mathcal{G}$ represents the average routing probability of ${E}_i$, which can be expressed by the following formulas:
\begin{equation}
\mathcal{F} = \frac{1}{K} \sum_{i=1}^{M} \bm{1}\{\arg\max \mathcal{P}(\bm{X}) = i\}, \; \mathcal{G} = \frac{1}{K} \sum_{i=1}^{K} \mathcal{P}(\bm{X})_i.
\label{eq:F_def}
\end{equation}
This distillation process encourages the synergy token to internalize both temporal motion cues and spatial geometry, forming a unified latent representation that supports downstream cross-vision reasoning:
\begin{equation}
    \mathcal{L} =  \mathcal{L}_{\text{coarse}} + \alpha \mathcal{L}_{\text{aux}}.
\end{equation}
It is worth noting that, in practice, content rarely exists simultaneously across all three modalities (image, video, and 3D).
Therefore, our training follows a progressive strategy: each modality is first pretrained independently, followed by partial joint training on paired video–image subsets.
Further implementation details are provided in Appendix~\ref{app:training}.

\begin{table*}[!t]
\centering
\fontsize{8}{9}\selectfont
\setlength{\tabcolsep}{1.5mm}
\begin{tabular}{lcccccc}
\toprule
\multirow{2}{*}{Model} & \multicolumn{3}{c}{ Image Understanding } & \multicolumn{3}{c}{Video Understanding} \\
\cmidrule(r){2-4}\cmidrule(r){5-7}
& MMStar~\cite{chen2024mmstar} & 3DSRBench$_\text{real}$~\cite{ma20253dsrbench} & MMSI-Bench~\cite{yang2025mmsibench} & VSI-Bench~\cite{yang2025vsibench} & CVBench
~\cite{zhu2025cvbench} & VideoMME$_\text{w/o sub.}$~\cite{fu2025videomme}  \\ 
\midrule
GPT-4o~\cite{hurst2024gpto} & 64.7  & 44.2 & 30.3 & 34.0 & 69.1 & 71.9 \\
Gemini-2.0-Flash~\cite{team2024gemini} & - &  49.8 & 69.4 & 45.4 & - & - \\
\hdashline
LLaVA-OV-7B~\cite{li2024llavaonevision} &  61.7 & 54.4 & 24.5 & 32.4 & 52.6 & 58.2\\
LLaVA-OV-1.5-8B~\cite{an2025llavaonversion15} & 67.7 & 57.8 & 29.6 & 46.3 & 21.7 & 56.3 \\
InternVL2.5-8B~\cite{chen2025internvl} & 62.8 & 50.9 & 28.7 & 41.6 & \bf 59.4 & 64.2\\
InternVL3-8B~\cite{zhu2025internvl3} & 68.2 & 58.1 & 25.7  & 50.7 & - &  66.3\\
Qwen2.5-VL-7B~\cite{bai2025qwen25vl} & 62.5 & 48.4 & 24.7 & 33.0 & 51.3 & 65.1\\
Qwen3-VL-8B~\cite{bai2025qwen25vl} & 55.0 &  59.9 & 29.0 & 34.8 & 50.2 & 52.1  \\
\hdashline
Spatial-MLLM-4B~\cite{wu2025spatialmllm} & 49.1 & 47.8 & 25.7 & 48.4 & 38.2 & 44.1 \\
SpaceR~\cite{ouyang2025spacer} & 53.6 & 57.4 & 27.2  & 24.9 & 50.4 & 56.9 \\
LLaVA-3D~\cite{zhu2025llava3d} & 31.7 & 39.3 & 20.8 & 9.36 & 33.2 & 26.5 \\
\hdashline
\rowcolor{lightblue} PolyV(Ours) & \bf 71.4\textcolor{red}{$_{\uparrow8.9}$} & \bf 63.4\textcolor{red}{$_{\uparrow15.0}$} & \bf 31.7\textcolor{red}{$_{\uparrow7.0}$} & \bf 52.7\textcolor{red}{$_{\uparrow19.7}$} & 59.1\textcolor{red}{$_{\uparrow7.8}$} & \bf 69.6\textcolor{red}{$_{\uparrow4.5}$}\\
\bottomrule
\end{tabular}
\vspace{-2mm}
\caption{Comparison of PolyV with existing MLLMs on image and video understanding benchmarks. All models are evaluated under the official benchmark metrics, and average scores are reported. Improvements over the backbone Qwen2.5-VL-7B~\cite{bai2025qwen25vl} are marked in \textcolor{red}{red}.}
\label{tab:image_video_understanding}
\vspace{-2mm}
\end{table*}

\begin{table*}[!t]
\centering
\fontsize{8}{9}\selectfont
\setlength{\tabcolsep}{2.0mm}
\begin{tabular}{lcccccccc}
\toprule
\multirow{2}{*}{Model} & \multicolumn{5}{c}{ ScanQA$_\text{val}$~\cite{azuma2022scanqa} } & \multicolumn{2}{c}{SQA3D$_\text{test}$~\cite{ma2022sqa3d}} & \multirow{2}{*}{Open-EQA$_\text{HM3D}$~\cite{majumdar2024openeqa}}\\
\cmidrule(r){2-6}\cmidrule(r){7-8}
& BLEU-1 & 	BLEU-4 & 	METEOR & 	ROUGE-L & 	CIDEr & 	EM-1 & 	EM-R1 & \\ 
\midrule
Qwen2.5-VL-7B~\cite{bai2025qwen25vl} &  27.8 &	3.0  &	11.4  &	29.3  &	53.9  &	46.5  &	49.8 & 56.6 \\
LLaVA-Video-7B~\cite{zhang2024videollava} & 39.7  &	3.1  &	17.7  &	44.6  & 88.7  &	48.5 &	 -	 & \\
\hdashline
Spatial-MLLM~\cite{wu2025spatialmllm} &	44.4 & 	14.8  &	18.4  &	45.0  &	91.8  &	 55.9  &	58.7 & - \\
Scene-LLM~\cite{fu2024scenellm} & - & 11.7 & 15.8 & 35.9 & 80.0 & 53.6 & - & -\\
LL3DA~\cite{chen2024ll3da} & - & - & 15.9 & 37.3 & 76.8 & - & - & -\\
LEO~\cite{huang2023embodied} & - & 11.5 & 16.2 & 39.3 & 80.0 & 50.0 & 52.4 & -\\
\hdashline
LLaVA-3D~\cite{zhu2025llava3d} &	- &	14.5 &	20.7 &	50.1 &	91.7 &	55.6 & 	57.6 & 53.2 \\	
Video-3D-LLM~\cite{zheng2025video3dllm} &	- &	16.4 &	20.0 &	49.3 &	102.1 &	58.6  &	–	& -\\
ROSS3D~\cite{wang2025ross3d} &	- &	17.9 &	20.9 &	50.7 &	107.0 &	63.0  &	65.7 & - \\
3DRS~\cite{huang20253drs} & 48.4 & 17.2 & 20.5 & 49.8 & 104.8 & - & - & -\\
Uni3D-MoE~\cite{zhang2025uni3dmoe} & 43.7 & 17.5 & 19.0 & 47.1 & 97.6 & 57.2 & - & -\\
\hdashline
\rowcolor{lightblue} PolyV(Ours) &	\bf 50.2\textcolor{red}{$_{\uparrow22.4}$} &	\bf 18.6\textcolor{red}{$_{\uparrow15.3}$} &	\bf 23.1\textcolor{red}{$_{\uparrow11.7}$} &	\bf 51.9\textcolor{red}{$_{\uparrow22.6}$} &	\bf 105.6\textcolor{red}{$_{\uparrow51.7}$} &	\bf 64.8\textcolor{red}{$_{\uparrow18.3}$}  &	\bf 67.5\textcolor{red}{$_{\uparrow17.7}$} & \bf 63.4\textcolor{red}{$_{\uparrow6.8}$} \\
\bottomrule
\end{tabular}
\vspace{-2mm}
\caption{Evaluation of 3D question-answering. General 2D VLMs~\cite{bai2025qwen25vl,zhang2024videollava} are evaluated in a zero-shot setting. }
\label{tab:3d_understanding}
\vspace{-2mm}
\end{table*}

\textbf{$\blacktriangleright$ Fine-grained Synergy Training.}
While the coarse-grained stage fosters general cross-vision synergy, it remains insufficient for modeling fine-grained synergy, particularly how objects and relations evolve across image, video, and 3D modalities.
To address this, we draw inspiration from~\cite{wu2025usg}, which represents object- and relation-level structures using scene graphs across modalities and design a fine-grained synergy training scheme that explicitly grounds reasoning at both the object level (e.g., spatial consistency, motion continuity) and the relation level (e.g., interaction dynamics and viewpoint-dependent changes).
As illustrated in Fig.~\ref{fig:training}(c), we construct a Cross-Vision Synergy Question–Answer (\textbf{CSQA}) dataset using GPT-4o~\cite{hurst2024gpt4o}, which automatically generates multimodal questions~\cite{gao2024generateanyscene} based on scene graphs to encourage reasoning about static structures and dynamic variations across modalities jointly, such as temporal consistency (image$\leftrightarrow$video) or spatial correspondence (image$\leftrightarrow$3D).
In total, we curate 20K fine-grained QA pairs, with construction details provided in Appendix~\ref{app:csqa_data}.
During training, multi-modal inputs and the corresponding synergy question are jointly fed into the model to generate a textual response, optimized using a combination of cross-entropy loss $\mathcal{L}{ce}$ over textual outputs and the expert load-balancing loss $\mathcal{L}{\text{aux}}$ introduced earlier.

\section{Experiments}

\subsection{Experimental Setups}

\textbf{Datasets.}
Our model is trained in multiple stages, each utilizing datasets tailored to the learning objectives of that stage, and detailed configurations are provided in Appendix~\ref{app:implementation}.
In brief, the primary data sources include image datasets, i.e., LLaVA 1.5-558K and LLaVA 1.5-Mix-665K~\cite{liu2024improved}; video datasets, i.e., LLaVA-Video-178K~\cite{zhang2024videollava} and ShareGPT4Video~\cite{chen2024sharegpt4video}; and 3D datasets, i.e., LLaVA-3D-Instruct-860K~\cite{zhu2025llava3d}.
During the fine-grained synergy training stage, we further incorporate 20K QA pairs constructed from the dataset in~\cite{wu2025usg}.

\textbf{Model Details.}
We initialize the LLM backbone with Qwen2.5-VL-7B~\cite{bai2025qwen25vl}, which supports both image and video inputs.
Following~\cite{zheng2025video3dllm}, we further incorporate 3D positional embeddings to enhance the model's spatial understanding.
In the second training stage, MoE and dense layers are interleaved at four-layer intervals (i.e., layers 0, 4, 8, 12, 16, 20, 24, and 28) to balance efficiency and capacity.
Each MoE layer consists of 4 experts, and the top-2 experts are dynamically activated per token during inference.

\textbf{Training Details.}
We adopt a two-stage training strategy as described in Sec.~\ref{sec:training}.
Across all stages, the models are optimized using the AdamW optimizer~\cite{loshchilov2017decoupled} in conjunction with a cosine learning rate scheduler.
A warm-up schedule with a warm-up ratio of 0.03 followed by cosine decay is applied independently for each stage.
The learning rate is set to $2\times10^{-5}$.
More details are provided in Appendix~\ref{app:training}.

\subsection{Main Results}

\paragraph{Image Understanding.}
We evaluate PolyV on image understanding tasks to assess its capability for both general and spatial reasoning from static inputs.
Specifically, we compare it with state-of-the-art VLMs on three benchmarks: MMStar~\cite{chen2024mmstar} for general understanding, and 3DSRBench~\cite{ma20253dsrbench} and MMSI-Bench~\cite{yang2025mmsibench} for spatial reasoning.
As shown in Table~\ref{tab:image_video_understanding}, PolyV consistently achieves the best overall performance across all benchmarks, outperforming its backbone Qwen2.5-VL-7B by about 10\% on average.
It also surpasses spatially specialized models (e.g., Spatial-MLLM~\cite{wu2025spatialmllm}, SpaceR~\cite{ouyang2025spacer}, LLaVA-3D~\cite{zhu2025llava3d}), achieving notable gains in spatial accuracy.
Notably, compared with SpaceR, PolyV under the same backbone exhibits clear gains in both spatial and general image understanding (53.6 vs. 71.4).
These results demonstrate that PolyV effectively captures synesthetic visual representations, enhancing spatial reasoning and generalization within the image modality.

\vspace{-2mm}
\paragraph{Video Understanding.}
We further evaluate PolyV on video-based benchmarks to examine its capability for temporal reasoning and cross-frame spatial understanding, including VideoMME~\cite{fu2025videomme} for general understanding, as well as on VSI-Bench~\cite{yang2025vsibench} and CVBench~\cite{zhu2025cvbench} for spatially grounded reasoning.
As shown in Table~\ref{tab:image_video_understanding}, PolyV achieves the best performance across all datasets.
In particular, on CVBench, which emphasizes spatial consistency and relational reasoning across frames, PolyV exhibits substantial gains over all baselines, underscoring its ability to transfer spatial and temporal priors across modalities effectively.

\begin{table}[!t]
\centering
\fontsize{8}{9}\selectfont
\setlength{\tabcolsep}{0.5mm}
\begin{tabular}{lcccc}
\toprule
 Setting    &  MMStar~\cite{chen2024mmstar} & Open-EQA~\cite{majumdar2024openeqa} &	VSI-Bench~\cite{yang2025vsibench}  & Video-MME~\cite{fu2025videomme} \\
 \midrule
    w/o expert & 68.9 & 60.3 & 45.8 & 66.4  \\
    w/ expert  &  71.4 & 63.4 & 52.7 & 69.6 \\
   \hdashline
   \rowcolor{lightgrey} \multicolumn{5}{l}{$\bullet$ \textit{Expert Numbers}} \\
   2  & 69.5 & 61.2 & 48.3 & 67.5 \\
   3  & 70.1 & 62.7 & 50.6 & 68.4 \\
   4 &   71.4 & 63.4 & 52.7 & 69.6 \\
   \hdashline
   \rowcolor{lightgrey} \multicolumn{5}{l}{$\bullet$ \textit{MoE layers}} \\
   First Half & 69.7 & 61.0 & 49.4 & 67.6 \\
   Last Half &   70.5 & 62.0 & 50.8 & 68.3 \\
   Interval(4) & 71.4 & 63.4 & 52.7 & 69.6\\
   Full & 70.2 & 61.8 & 53.5 & 68.1 \\
\bottomrule
\end{tabular}
\vspace{-2mm}
\caption{Ablation Studies on the MoE design in PolyV, examining its necessity, expert number, and architectural placement.}
\label{tab:moe_ablation}
\vspace{-3mm}
\end{table}

\begin{table}[!t]
\centering
\fontsize{8}{9}\selectfont
\setlength{\tabcolsep}{0.5mm}
\begin{tabular}{lcccc}
\toprule
 Setting    &  Open-EQA &	3DSRBench$_\text{real}$  &	VSI-Bench  & Video-MME \\
 \midrule
 PolyV(Full) & 63.4 & 63.4 & 52.7 & 69.6\\
 \hdashline
 \rowcolor{lightgrey} \multicolumn{5}{l}{$\bullet$ \textit{Training Strategy}} \\
  only Coarse-grained  & 61.9 & {61.0} & 51.6 & 68.2 \\
   only Fine-grained & {62.5} & 59.8 & {50.2} & {68.9} \\
   \hdashline
   \rowcolor{lightgrey} \multicolumn{5}{l}{$\bullet$ \textit{Foundation Model}} \\
    w/ VideoFM  & 62.8 & 61.4 & {52.1} & {69.3} \\
   w/ 3DFM & {63.0} & {62.7} & 51.5 & 68.8 \\
\bottomrule
\end{tabular}
\vspace{-2mm}
\caption{Comparison studies on the training strategies and foundation model utilization in PolyV.}
\label{tab:training_strategy}
\vspace{-3mm}
\end{table}

\vspace{-2mm}
\paragraph{3D Understanding.}
We evaluate PolyV on 3D scene understanding tasks using three representative benchmarks: ScanQA~\cite{azuma2022scanqa}, SQA3D~\cite{ma2022sqa3d}, and Open-EQA~\cite{majumdar2024openeqa}.
Following~\cite{huang20253drs,zhang2025uni3dmoe}, we report CIDEr, BLEU-1, BLEU-4, METEOR, and ROUGE-L for ScanQA, adopt Exact Match (EM) and Refined Exact Match (EM-R) for the SQA3D test set, and LLM-Match score~\cite{majumdar2024openeqa} for Open-EQA.
As shown in Table~\ref{tab:3d_understanding}, PolyV consistently outperforms its backbone and prior 3D reasoning models, including video-based VLMs and 3DRS~\cite{huang20253drs}.
Such observation indicates that the cross-vision synergy learned from image and video modalities effectively transfers to 3D, enhancing geometric reasoning and holistic spatial understanding.

\begin{figure}[!t]
    \centering
    \includegraphics[width=0.99\linewidth]{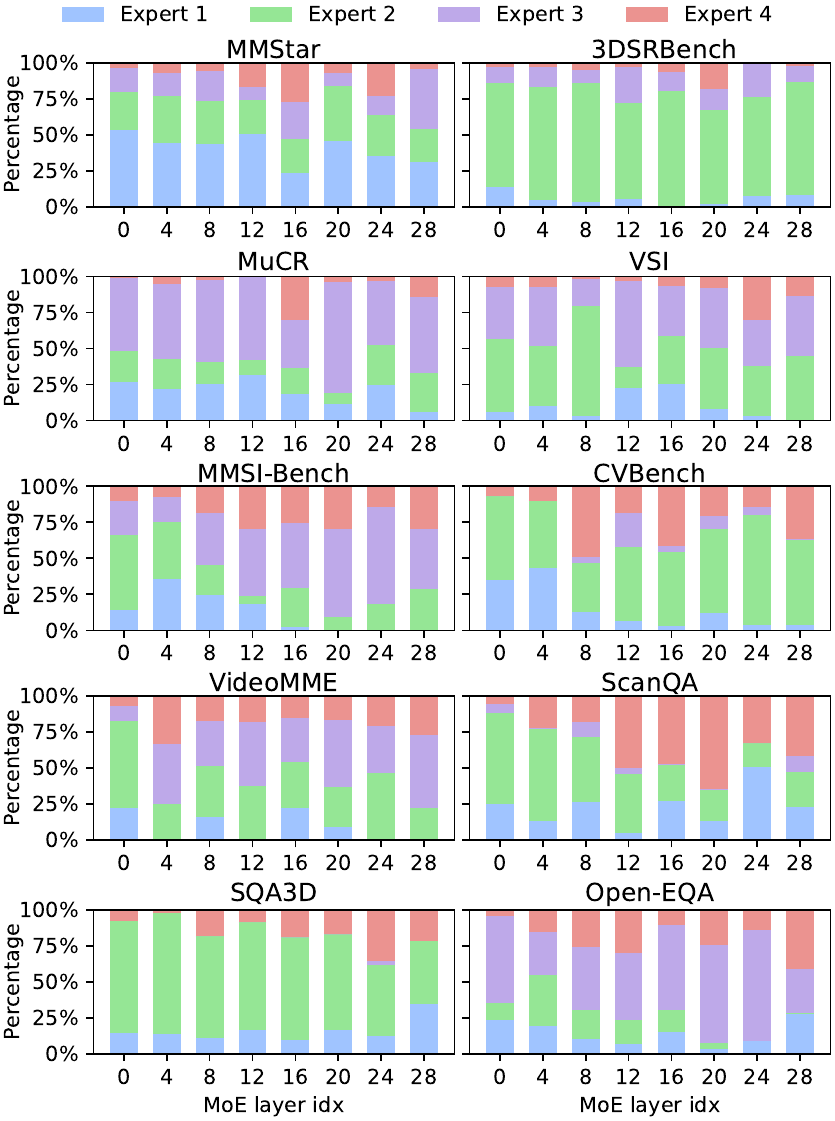}
    \vspace{-2mm}
    \caption{Token distribution across experts on different benchmarks, illustrating load balance and routing diversity that reflect PolyV's adaptive expert specialization.}
    \label{fig:benchmark_expert}
    \vspace{-2mm}
\end{figure}

\begin{figure*}
    \centering
    \includegraphics[width=0.90\linewidth]{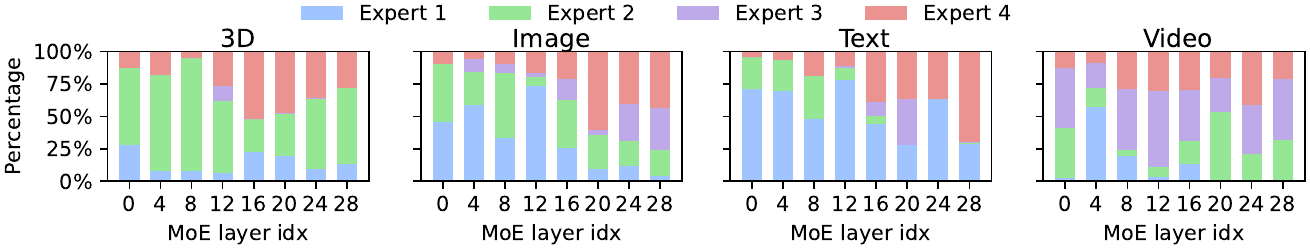}
    \vspace{-2mm}
    \caption{Token distribution across experts by modality. “Percentage” denotes the token fraction per expert across all MoE layers.}
    \label{fig:modality_expert}
    \vspace{-2mm}
\end{figure*}

\begin{figure*}[!t]
    \centering
    \includegraphics[width=0.99\linewidth]{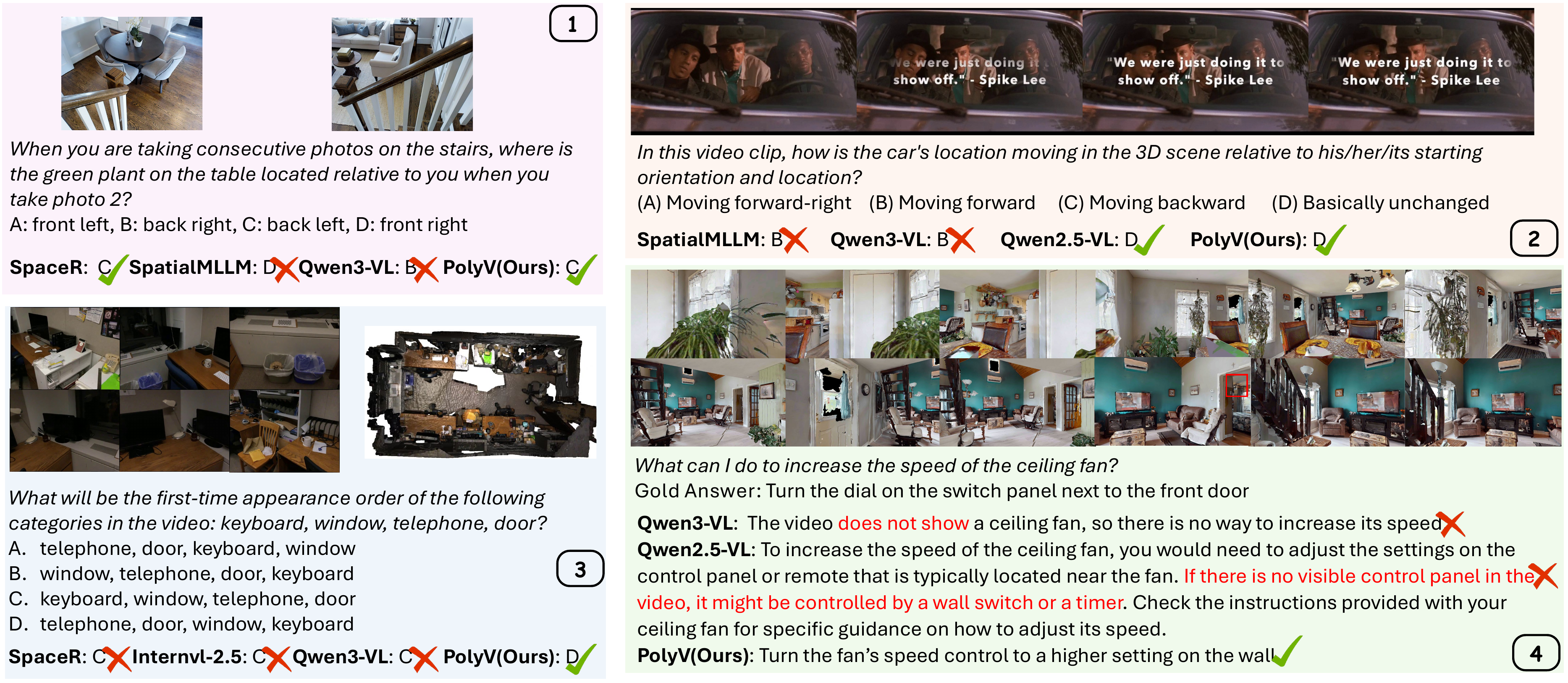}
    \vspace{-2mm}
    \caption{Qualitative comparisons between PolyV and existing models.
Case 1 is from MMSI-Bench~\cite{yang2025mmsibench}, Case 2 from DSI-Bench~\cite{zhang2025dsibench}, Case 3 from VSI-Bench~\cite{yang2025vsibench}, and Case 4 from Open-EQA (HM3D)~\cite{majumdar2024openeqa}. 
}
    \label{fig:case_study}
    \vspace{-2mm}
\end{figure*}

\subsection{In-Depth Analyses and Discussion}

\paragraph{Comparative Analysis of Uni-MoE and Dense Models.}
We compare dense models (without experts) and PolyV (with expert-based Uni-MoE architecture) under identical training settings, differing only in architectural design.
As shown in Table~\ref{tab:moe_ablation}, PolyV yields superior results over the dense variant on all benchmarks, demonstrating the MoE framework’s superiority in enabling more efficient specialization and cross-modal knowledge sharing than uniform dense parameterization, to enhance representational capacity and adaptive reasoning.

\vspace{-2mm}
\paragraph{Effect of the Number of Experts.}
We investigate the impact of varying the number of experts within MoE
layers while keeping the number of activated
experts the same, as detailed in Table~\ref{tab:moe_ablation}.
Increasing the total number of experts consistently improves performance, with sparse configurations surpassing the single-expert dense model by 2.5\% on MMStar, 3.1\% on Open-EQA, 6.9\% on VSI-Bench, and 3.2\% on VideoMME.
To balance efficacy and efficiency, we choose 4 experts in our setting.

\vspace{-2mm}
\paragraph{Analyzing the Architectures of PolyV.}
Table~\ref{tab:moe_ablation} compares four MoE configurations: ``First-Half'' applies MoE layers to the lower half of the network, ``Second-Half'' to the upper half, ``Interval(4)'' interleaves MoE and dense layers every four layers, and the ``Full'' converts all layers to sparse MoE layers. 
We observe that the ``Full'' setup yields no clear performance gain while increasing training cost, whereas ``Interval(4)'' achieves the best balance between efficiency and representational specialization.

\vspace{-2mm}
\paragraph{Effect of Training Strategy.}
We compare three training configurations: the coarse-grained strategy, the fine-grained strategy, and their combination.
As shown in Table~\ref{tab:training_strategy}, coarse-grained tuning excels on 3D benchmarks (e.g., 3DSRBench), highlighting its strength in capturing global structural representations.
Fine-grained tuning, in contrast, focuses on detailed reasoning and performs better on video-related tasks (e.g., VSI-Bench, VideoMME), though with slightly weaker spatial performance.
Combining both yields the best overall results, confirming that coarse structural alignment and fine object-/relation-level reasoning are complementary for achieving robust cross-vision synergy.

\vspace{-2mm}
\paragraph{Effect of Foundation Model Strategy.}
We evaluate the impact of integrating video and 3D foundation models into PolyV.
As shown in Table~\ref{tab:training_strategy}, integrating VideoFM notably improves performance on video benchmarks (VSI-Bench and VideoMME), reflecting stronger temporal reasoning, while 3DFM improves results on spatial tasks (3DSRBench), indicating better geometric understanding.
These results show that temporal and spatial priors distilled from strong single-modality foundation models provide complementary guidance for achieving robust cross-vision synergy.

\vspace{-2mm}
\paragraph{Routing Distributions.}
Fig.~\ref{fig:benchmark_expert} illustrates the routing distributions of the four experts across different benchmarks.
Expert 1 is primarily activated in general image understanding (e.g., MMStar), while Expert 2 dominates spatial reasoning (3DSRBench, SQA3D). 
Expert 3 is primarily triggered in temporal and motion-focused datasets (MuCR, VideoMME), and Expert 4 shows balanced activation across mixed or reasoning tasks (MMSI-Bench, CVBench). 
These patterns confirm that PolyV adaptively allocates experts according to task priors, enabling effective cross-vision synergy.
Furthermore, we also visualize the token distribution of the four experts across different modalities, as demonstrated in Fig.~\ref{fig:modality_expert}.

\vspace{-2mm}
\paragraph{Case Study.}
Fig.~\ref{fig:case_study} presents qualitative examples across diverse visual reasoning tasks.
In Cases 1 and 3, PolyV accurately interprets object orientation and viewpoint relations in image-based and multi-view spatial reasoning.
In Case 2, focusing on video temporal and spatial understanding, it correctly infers that no spatial state changes occur.
Finally, in Case 4, PolyV effectively identifies key spatial cues in 3D environments, producing coherent responses.
These examples collectively highlight PolyV's ability to dynamically leverage multimodal representations for contextually adaptive and fine-grained visual reasoning.

\section{Conclusion}
In this work, we introduced \textbf{PolyV}, a unified LVM designed to achieve synergy across image, video, and 3D modalities. 
Technically, we proposed a MoE architecture coordinated by a dynamic router to enable adaptive modal-specific\&sharing feature interaction. 
We further develop a synergy-aware training framework that integrates coarse-grained cross-modal knowledge distillation and fine-grained cross-vision alignment, enabling PolyV to capture both global and object/relation-level correspondences across modalities.
Extensive experiments on diverse benchmarks demonstrate that PolyV surpasses existing LVMs in spatial, temporal, and cross-modal reasoning, paving the way toward human-like cross-vision synergy learning.

{
    \small
    \bibliographystyle{ieeenat_fullname}
    \bibliography{main}
}

\clearpage
\appendix

\makeatletter
\let\addcontentsline\origaddcontentsline
\makeatother

\tableofcontents

\section{Extended Model Architecture}
\label{app:}

\paragraph{Vision Encoder}
The universal vision encoder is built upon the architecture proposed in~\cite{bai2025qwen25vl}, which adopts a redesigned Vision Transformer (ViT) backbone.
Specifically, the height and width of the input images are resized to multiples of 28 before being fed into the ViT.
Then, the images are split into patches with a stride of 14. 
For video data, every two consecutive frames are grouped into a single unit to reduce the number of visual tokens and alleviate computational overhead.
To address the quadratic complexity of standard self-attention when processing images of varying resolutions, a windowed attention with a maximum window size of 112×112 (corresponding to 8×8 patches) is introduced in most layers, ensuring that computational cost scales linearly with the number of patches rather than quadratically.
Additionally, the positional encoding scheme is extended from 2D to 3D patch partitioning, enabling the encoder to jointly model both spatial and temporal dimensions for unified image and video understanding.
To inject 3D information into conventional 2D video frames, following~\cite{zheng2025video3dllm,wang2025ross3d}, we first calculate a set of global coordinates $(x, y, z)$ of each pixel at the position:
\begin{equation}
    [x\; y\; z\; 1] = [\textbf{D}_{i,j}\; \cdot\; [j\; i\; 1] \; \cdot (\textbf{K}^{-1})^{\top}\; 1 ]\;\cdot \textbf{B}^{\top},
\end{equation}
where $\textbf{D} \in \mathbb{R}^{H \times W}$ denotes the depth maps, $\textbf{B} \in \mathbb{R}^{4 \times 4}$ the extrinsic matrix, and $\textbf{K} \in \mathbb{R}^{3 \times 3}$ the camera intrinsic matrix.
A sinusoidal positional encoding is then applied to these 3D coordinates to obtain coordinate embeddings, which are subsequently added to the visual token embeddings, forming position-aware representations that effectively capture spatial and temporal structure.

\paragraph{MoE Layer.}
Each MoE layers consists of 4 parallel expert modules, implmented as the FFN-styled architecture in the LLM. 
These experts expand the feature dimension from  3584 to an intermediate dimension of 18944 via parallel gated projections (gate\_proj and up\_proj), followed by SiLU activations and dimensional reduction back to 3584 (down\_proj).
Tokens are dynamically routed to these experts using a learnable Top-K gating network (i.e., a linear layer), which adaptively selects the most suitable experts based on token-level semantic characteristics. In our implementation, we set K to 2.

\paragraph{Video Foundation Model}
In our work, we employ V-JEPA 2~\cite{assran2025vjepa2} as the video foundation Model. 
Architecturally, the model is parameterized as a vision transformer~\cite{} with a tubelet size of $2 \times 16 \times 16$ ($T \times H \times W$) to patchify the input video frames.
A 3D-RoPE is adopted to encode relative position information in the vision transformer. 
V-JEPA 2 has been pre-trained on more than 1 million hours of video via a self-supervised manner to comprehensively capture the broad temporal motion and action within the video.

\paragraph{3D Foundation Model}
This work employs the VGGT~\cite{wang2025vggt} as the 3D foundation teacher model to distill 3D geometric information, such as inter-frame correspondences within input frames.
The VGGT is built based on a fairly standard large transformer, with an alternating-attention design. 
Specifically, the frame-wise self-attention attends to the tokens within each frame separately, and global self-attention attends to the tokens across all frames jointly.
The alternating attention is devised to make the transformer focus within each frame and globally alternately. 
During training, the VGGT is optimized to predict a full set of 3D attributes, including camera parameters, depth maps, point maps, and 3D point tracks.
Recent works~\cite{wang2025vggt,chen2025thinkwith3d,zheng2025learning,lin2025evo} have shown that the VGGT can serve as a 3D genmetric feature extractor to enhance the downstream tasks.

\paragraph{Alignment Layer for Coarse-grained Synergtic Learning.}
To ensure dimensional compatibility between the generated synergistic latent features and the modality-specific priors extracted from video and 3D foundation models, we introduce dedicated alignment layers that project the synergy representations into the corresponding feature spaces.
These alignment modules are intentionally designed with slight architectural differences, reflecting the distinct dimension nature of temporal features in videos and geometric features in 3D scenes.

\paragraph{}

\section{More Implementation Details}
\label{app:implementation}

\begin{figure*}[!t]
    \centering
    \includegraphics[width=0.99\linewidth]{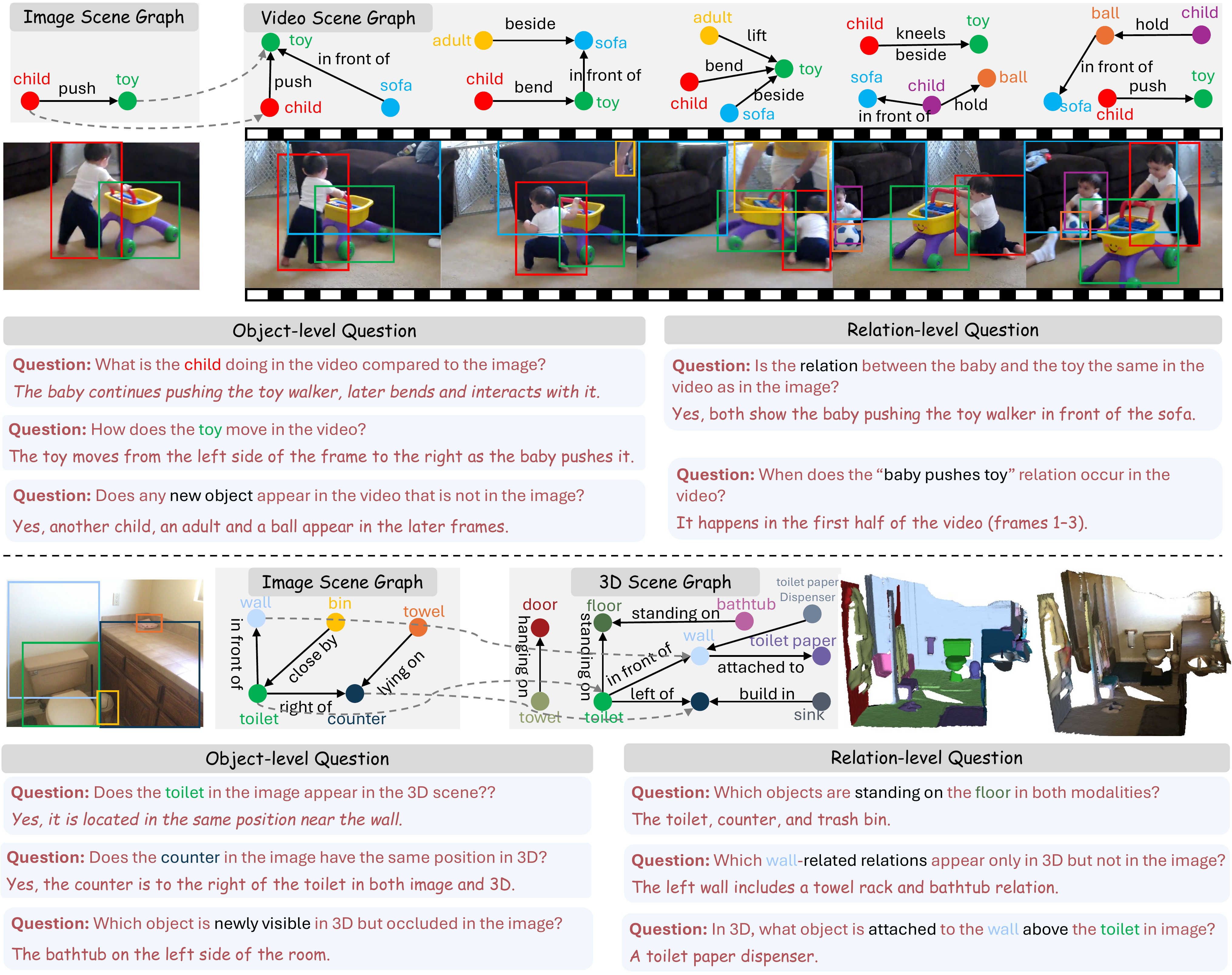}
    \vspace{-2mm}
    \caption{Illustration of CSQA examples, which is constructed based on a paired Image-Video scene graph and Image-3D scene graph.}
    \label{fig:CSQA}
\end{figure*}

\subsection{CSQA Data Construction}
\label{app:csqa_data}
Here, we show the detailed cross-vision synergy question-answer (CSQA) pair construction pipeline.

\paragraph{Data Source.}
We construct the CSQA dataset using paired scene-graph annotations from image–video and image–3D sources, derived from two established multimodal SG datasets~\cite{wu2025usg}:
\begin{compactitem}
    \item \textbf{Image-Video Scene Graph} is constructed based on PVSG~\cite{yang2022panoptic}, AG~
\cite{ji2020actiongenome}, where the first frame of each video serves as the image view and non-adjacent clips provide the corresponding video segments. Cross-modal object associations are directly taken from the existing annotations, as follows.
    \item \textbf{Image-3D Scene Graph} is derived from 3DSG~\cite{armeni20193d} by sampling 2D views linked to 3D scenes. Furthermore, the 3D objects are ground-annotated into the sampled images, and relationship annotations are inherited to form complete paired SGs.
\end{compactitem}

\paragraph{QA Generation.}
To capture fine-grained cross-vision synergy, we design two categories of questions: \textbf{object-level} (e.g., spatial consistency, motion continuity) and \textbf{relation-level} (e.g., interaction dynamics, viewpoint-dependent changes), as illustrated in Fig.~\ref{fig:CSQA}.
Given the paired scene graphs, we prompt GPT-4o~\cite{hurst2024gpt4o} to generate QA pairs targeting these specific aspects.
For quality control, each generated question is validated using Qwen25-VL-72B~\cite{bai2025qwen25vl}, which answers the question based on both the scene graph and the corresponding visual inputs.
Only QA pairs whose predicted answers match the gold responses are retained; mismatched samples are discarded to ensure reliability of the final CSQA dataset.
Finally, we obtain the 20K CSQAs, which contain diversified questions and a focus.
\begin{tcolorbox}[breakable, fontupper=\customfont, title=Prompt for Object-level CSQA]
\vspace{-2mm}
{\small
\textbf{Instruction}: 
You are given paired scene graphs, one from an image and one from a video. Your task is to generate object-level, cross-vision question–answer pairs that explicitly compare or link the image to the video.
Your generated QAs must rely only on the information encoded in the scene graphs. \\
Each QA should:
    \begin{compactitem}
        \item (1) Be object-focused: Ask about a specific object (e.g., child, toy, sofa, ball); Not general high-level scene understanding; Not reasoning beyond objects \& relations
        \item (2) Be cross-vision: Questions must compare the image with the video appearance or disappearance of an object; change of position; object motion; object interaction; new objects entering the scene; object state change.
        \item (3) Be explicitly grounded. The question must be answerable only using the provided scene graph information.
        \item (4) Be natural, concise, human-like
    \end{compactitem}
Important Constraints:
\begin{compactitem}
    \item Do NOT hallucinate objects, attributes, or relations not found in the scene graphs.
    \item Avoid global reasoning (e.g., “What happens in the scene?”).
    \item Do NOT describe entire events; stay object-focused.
    \item Each question must be answerable and concise.
    \item Each answer must be grounded strictly in the paired SG content.
    \item Output QAs must be in JSON format.
\end{compactitem}
====================\\
Examples (DO NOT COPY; FOLLOW SIMILAR STYLE):\\
\{"question": "What is the child doing in the video compared to the image?", "answer": "The baby continues pushing the toy walker, later tends and interacts with it."\},\\
\{"question": "How does the toy in the image move in the video?", "answer": "The toy moves from the left side of the frame to the right as the baby pushes it."\}\\
\{"question": "Does any new object appear in the video that is not in the image?", "answer": "Yes, another adult and a ball appear in the later frames. "\}\\
==================\\
\textbf{Input Data}: [\textit{paired image-3D scene graph}] \\
\textbf{Output QAs}:
}
\end{tcolorbox}

\begin{tcolorbox}[breakable, fontupper=\customfont, title=Prompt for Relation-level CSQA]
\vspace{-2mm}
{\small
\textbf{Instruction}: 
You are given a paired Image Scene Graph (ISG) and Video Scene Graph (VSG).
Your task is to generate relation-level, cross-vision, synergy-focused question–answer pairs, comparing relational changes between the image and the video. \\
Each QA must:
\begin{compactitem}
    \item be relation-centered, i.e., focusing on pairwise or ternary relations
    \item Connect image and video explicitly. Questions must compare: whether a relation persists or changes; when a relation appears / disappears; how a relation evolves over time; whether a relation becomes stronger/weaker or switches type; whether a new relation emerges only in the video.
    \item Stay grounded in the scene graphs, no hallucinated objects or relations.
    \item Produce natural, human-like questions
\end{compactitem}
Important Constraints:
\begin{compactitem}
    \item Only use relations that appear explicitly in the scene graphs.
    \item No event hallucination.
    \item Keep questions short, clear, and relation-focused.
    \item Answers must be concise and grounded (no speculation).
    \item Avoid high-level reasoning beyond relations.
    \item Output Q\&As must be in JSON format.
\end{compactitem}
====================\\
Examples (DO NOT COPY; FOLLOW SIMILAR STYLE):\\
\{"question": "Is the relation between the baby and the toy the same in the video as in the image?", "answer": "Yes, both show the baby pushing the toy walker in front of the sofa."\},\\
\{"question": "Does any new relation appear in the video that is not present in the image?", "answer": "Yes. The man and another child appear in front of the sofa."\}\\
====================\\
\textbf{Input Data}: textual captions \\
\textbf{Output QAs}:
}
\end{tcolorbox}

\subsection{Model Details.}

We initialize the LLM backbone with Qwen2.5-VL-7B~\cite{bai2025qwen25vl}, which supports both image and video inputs.
Following~\cite{zheng2025video3dllm}, we further incorporate 3D positional embeddings to enhance the model's spatial understanding.
In the second training stage, MoE and dense layers are interleaved at four-layer intervals (i.e., layers 0, 4, 8, 12, 16, 20, 24, and 28) to balance efficiency and capacity.
Each MoE layer consists of 4 experts, and the top-2 experts with the highest routing probabilities are dynamically activated for each token during inference.
We show the detailed architecture and parameters number in Table~\ref{tab:model_architecture}.

\begin{table*}[!t]
    \centering
    \caption{Detailed Architecture of PolyV and Qwen2.5-VL(Base). ``Width'' represents the dimension of the hidden states. ``FFN'' denotes the dimension of the feed-forward network’s intermediate layer. ``FFN Factor'' represents the quantity of linear layers in the FFN. ``Activated'' or ``Total Param'' refers to the activated or total number of parameters. We highlight the architecture of PolyV in \colorbox{lightblue}{blue}.}
    \label{tab:model_architecture}
    \vspace{-2mm}
    \fontsize{9}{11}\selectfont
\setlength{\tabcolsep}{2.5mm}
    \begin{tabular}{lccccccccccc}
    \toprule
       Name  & Experts & Top-k & \makecell{MoE\\ Layers} & Embedding & Width & Layers & FFN & \makecell{FFN\\ Factor} & Heads & \makecell{Activated\\  Param.} & \makecell{Total\\ Param.} \\
       
    \midrule
      \rowcolor{lightgrey} Base   & - & - & - & 152064 & 3584 & 28 & 18944 & 3 & 28 & 7.0B & 7.0B\\
      \hdashline
      First Half & 4 & 2 & 7 & 152064 & 3584 & 28 & 18944 & 3 & 28 & 8.4B & 11.3B\\
      Last Half & 4 & 2 & 7 & 152064 & 3584 & 28 & 18944 & 3 & 28 & 8.4B & 11.3B \\
     \rowcolor{lightblue} Interval(4) & 4 & 2 & 7 & 152064 & 3584 & 28 & 18944 & 3 & 28 & 8.4B & 11.3B \\
      Full & 4 & 2 & 14 & 152064 & 3584  & 28 & 18944 & 3 & 28 & 9.9B & 15.6B \\
    \bottomrule
    \end{tabular}
\end{table*}

\begin{table*}[!t]
\centering
\caption{The hyperparameter settings in each stage.}
\label{tab:hyperparameters}
\vspace{-2mm}
\fontsize{9}{11}\selectfont
\setlength{\tabcolsep}{2.5mm}
\begin{tabular}{lcccc}
\toprule
Parameters  &  Stage-1-1 &  Stage-1-2  &  Stage-2-1 &  Stage-2-2 \\
\midrule
Optimizer & AdamW & AdamW & AdamW & AdamW  \\
Scheduler &  Cosine & Cosine & Cosine & Constant \\
Learning rate   & 2e-5 & 2e-6 & 1e-4 & 1e-5\\
Weight Decay & 0.05  & 0.05 & 0.05 & 0.1 \\
Warmup Ratio   & 0.03 & 0.03 & 0.05 & 0.05 \\
Updated Module & Projector & FFN & \makecell{Router\\ FFN (Experts) \\ Alignment } &  \makecell{Router\\ FFN (Experts)} \\
Data Num. & 40K & 45K & 50K & 40K\\
Data Source & LLaVA-3D-Instruct-860K &  \makecell{LLaVA 1.5-558K \\ LLaVA-Video-178K \\ LLaVA-3D-Instruct-860K } & \makecell{ShareGPT4Video \\ 3RScan \\ Scan2Cap}  & \makecell{CSQA \\ LLaVA 1.5-Mix-665K \\ LLaVA-Video-178K \\  LLaVA-3D-Instruct-860K}\\
\bottomrule
\end{tabular}
\end{table*}

\subsection{Training Details}
\label{app:training}

We provide additional details of the full training pipeline in this section. The complete set of hyperparameter configurations and training settings is summarized in Table~\ref{tab:hyperparameters}.

\paragraph{Stage-1-1: Alignment Pre-training }
In this stage, we establish the initial alignment between the vision encoder and the backbone LLM. Since our model is initialized from Qwen2.5-VL-7B, which already supports image and video understanding, we focus exclusively on optimizing the model with 3D data to equip it with an initial capability for 3D comprehension.
During this process, only the parameters of the projector are updated, while all other components remain frozen.

\paragraph{Stage-1-2: Instruction-Following Pre-training}
In this stage, we train separate model variants on modality-specific data, allowing the model to acquire specialized capabilities, such as temporal dynamics for videos and spatial geometry for 3D scenes, using richer vision–language instruction corpora.
During this process, only the feed-forward network (FFN) parameters are updated, while all other components remain frozen.
This design allows modality-specific expertise to be acquired efficiently without interfering with the shared cross-modal representations established in the previous stage.

\paragraph{Stage-2-1: Coarse-grained Synesthetic Learning}
This stage aims to equip the model with cross-vision synergy, enabling synesthetic reasoning across different visual modalities. Architecturally, the modality-specialized FFNs obtained from Stage-1-2 are used to initialize the MoE layers.
During training, we first freeze all experts and update only the router and alignment layers to stabilize cross-modal routing behavior. Once the routing becomes stable, we subsequently unfreeze the experts and jointly optimize the experts, router, and alignment layers to achieve coarse-grained synesthetic fusion across modalities.

\paragraph{Stage-2-2: Fine-grained Synesthetic Learning}
In this stage, the model is further fine-tuned using the constructed CSQA dataset to enhance its fine-grained cross-vision synesthetic reasoning capabilities. To maintain the model’s generality, we interleave a proportion of the modality-specific instruction data used in Stage~1-2 during training.
Consequently, this stage primarily updates the MoE layers, enabling more precise and discriminative synergistic interactions across modalities.

\paragraph{Overall Parameters settings.}
Across all stages, the models are optimized using the AdamW optimizer~\cite{loshchilov2017decoupled} in conjunction with a cosine learning rate scheduler.
A warm-up schedule with a warm-up ratio of 0.03 followed by cosine/constant decay is applied independently for each stage.
To enhance computational efficiency and reduce memory usage, we employ BF16 mixed-precision training.
For video and 3D multi-view inputs, the maximum number of frames is set to 32 with 640$\times$480 resolution.

\subsection{Benchmarks}

We detail the benchmarks we employed in our evaluation.

\paragraph{MMStar~\cite{chen2024mmstar}.} 
MMStar is a large-scale benchmark designed to evaluate truly vision-indispensable capabilities of multimodal models. The dataset contains 1,500 carefully curated human-verified samples across six core abilities and eighteen sub-dimensions. It mitigates two major issues in existing benchmarks—visual redundancy and potential data leakage, by ensuring that each question fundamentally requires visual understanding.

\paragraph{3DSRBench~\cite{ma20253dsrbench}.}
3DSRBench focuses on 3D spatial reasoning, assessing whether models can understand object height, orientation, position, inter-object relations, and multi-view consistency under both common and uncommon viewpoints. The benchmark consists of 2,772 manually annotated VQA samples, including approximately 2,100 real-world images and 672 synthetic multi-view images.

\paragraph{MMSI-Bench~\cite{yang2025mmsibench}.}
MMSI-Bench evaluates multi-image spatial intelligence, requiring models to integrate information across multiple views rather than reasoning from a single image. The benchmark comprises 1,000 carefully constructed multiple-choice questions sourced from over 120,000 candidate images, each accompanied by human-designed step-by-step reasoning. The benchmark includes a taxonomy that covers grounding, scene reconstruction, contextual transformation, and spatial logic failures.

\begin{table*}[!ht]
\centering
\caption{Detailed evaluation results on \textbf{MMStar}~\cite{chen2024mmstar}. Best results are marked in \textbf{bold}.}
\label{tab:mmstar}
\vspace{-2mm}
\fontsize{9}{11}\selectfont
\setlength{\tabcolsep}{3.0mm}
\begin{tabular}{lccccccc}
\toprule
Model & \makecell{Coarse \\ Perception} & \makecell{Fine-grained \\ Perception} & \makecell{Instance \\ Reasoning} & \makecell{Logical \\ Reasoning} & \makecell{Math} & \makecell{Science \\Technology} & Average \\
\midrule
LLaVA-NeXT-Video-7B~\cite{zhang2024llavanext-video} & 52.4 & 	25.6 & 	41.6 & 	28.0 & 	29.2 & 	25.2  & 33.7 \\
LLaVA-OneVision-7B~\cite{li2024llavaonevision} & 65.2 & 	51.2 & 	64.8 & 	58.0 & 	53.2 & 	44.4 & 56.1 \\
LLaVA-OneVision1.5-8B~\cite{an2025llavaonversion15} & 68.4 & 	53.6 & 	70.0  & 	70.8 & 	\bf 77.6 & 	\bf 59.2 & 66.6 \\
InternVL3-8B~\cite{chen2025internvl} & 74.0 & 	59.2 & 	72.0 & 	65.6 & 	64.4 & 	54.4 & 64.9\\
Qwen3-VL-7B~\cite{bai2025qwen25vl} & 75.2 & 58.0 & 69.2 & 51.2 & 34.8 & 41.6 & 55.0 \\
\hdashline
SpaceR~\cite{ouyang2025spacer} & 66.0 & 48.4 & 68.0 & 56.0 & 44.8 & 38.4 & 53.6 \\
SpatialMLLM~\cite{wu2025spatialmllm} & 63.6 & 43.2 &  55.6 & 47.6 & 54.0 & 30.8 & 49.1 \\ 
LLaVA3D~\cite{zhu2025llava3d} & 54.8 & 22.8 & 41.6 & 28.0 & 19.2 & 24.0 & 31.7 \\
Video3D-LLM~\cite{zheng2025video3dllm} & 23.6 & 3.2 & 18.8 & 11.6 & 14.8 & 18.0 & 15.0 \\
Ross3D~\cite{wang2025ross3d} & 41.2 & 10.0 & 30.8 & 20.0 & 12.0 & 16.0 & 21.7 \\
\hdashline
\rowcolor{lightblue} PloyV (Ours) &  \bf 82.0 & \bf 70.0 & \bf 78.0 & \bf 74.0 & 68.0 & 57.0 & \bf 71.4 \\
\bottomrule
\end{tabular}
\end{table*}

\begin{table}[!ht]
\centering
\caption{Detailed evaluation results on \textbf{3DSRBench}$_\text{real}$~\cite{ma20253dsrbench}. `Loc.' denotes location, `Orient.' means orientation, and `Multi.' is multi-object.}
\label{tab:3dsrnench}
\vspace{-2mm}
\fontsize{9}{11}\selectfont
\setlength{\tabcolsep}{0.5mm}
\begin{tabular}{lccccc}
\toprule
Model   & Overall &  Height &  Loc. &  Orient. &  Multi. \\
\midrule
LLaVA-v1.5-7B~\cite{liu2024improved} &  38.1 &  39.1 &  46.9 &  28.7 &  34.7 \\
LLaVA-NeXT-Video-7B~\cite{zhang2024llavanext-video} & 49.4 & 52.6 & 55.3 & 42.6 & 48.5  \\
LLaVA-OneVision-7B~\cite{li2024llavaonevision} & 54.4 & 56.8 & 61.3 & 46.1 & 50.3 \\
LLaVA-OneVision1.5-8B~\cite{an2025llavaonversion15} & 57.8 & 57.4 & 66.5 & 49.9 & 53.4  \\
InternVL2.5-8B~\cite{chen2025internvl}  & 50.9 &  45.9 &  68.1 &  38.7 &  43.3 \\
InternVL3-8B~\cite{zhu2025internvl3} & 58.1 & \bf61.3 & 67.6 & 48.3 & 53.8 \\
QWen2.5-VL-7B~\cite{bai2025qwen25vl} & 48.4 &  44.1 &  62.7 &  40.6 &  40.5  \\
Qwen3-VL-8B~\cite{bai2025qwen25vl} &  60.0 & 53.8 & 74.0 & 51.5 & 53.6 \\
\hdashline
LLaVA3D~\cite{zhu2025llava3d} & 39.3 & 49.1 & 36.8 & 25.0 & 46.3 \\
Ross3D~\cite{wang2025ross3d} & 39.6 & 36.5 & 40.2 & 35.9 & 42.4 \\
SpaceR~\cite{ouyang2025spacer} & 57.4 & 51.5 & 69.7 & 50.1 & 51.9 \\
SpatialMLLM~\cite{wu2025spatialmllm} & 47.8 & 49.6 & 49.0 & 42.4 & 49.0 \\
\rowcolor{lightblue} PloyV (Ours) & \bf 63.4 & 60.0 & \bf78.0 & \bf 57.0 & \bf58.0 \\
\hdashline
\bottomrule
\end{tabular}
\end{table}

\paragraph{CV-Bench~\cite{tong2024cambrian}.}
CV-Bench offers a vision-centric evaluation suite designed for 2D and 3D understanding in multimodal models. It contains 2,638 human-validated samples built upon datasets such as ADE20K~\cite{zhou2017scene}, COCO~\cite{lin2014microsoft}, and Omni3D~\cite{brazil2023omni3d}. The benchmark focuses on determining whether LVLMs truly rely on visual information rather than linguistic priors, emphasizing core perception capabilities such as localization, recognition, and spatial reasoning.

\paragraph{VSI-Bench~\cite{yang2025vsibench}.}
VSI-Bench evaluates visual spatial intelligence from videos, extending spatial understanding beyond static imagery. 
It includes over 5,000 QA pairs derived from 288 real videos.
These videos are sourced from the validation sets of the public indoor 3D scene reconstruction datasets ScanNet~\cite{dai2017scannet}, ScanNet++~\cite{yeshwanth2023scannet++}, and ARKitScenes~\cite{baruch2021arkitscenes}. 
Tasks encompass configurational reasoning, metric estimation, and spatiotemporal reasoning, enabling a comprehensive assessment of spatial understanding.

\paragraph{Video-MME~\cite{fu2025videomme}.}
Video-MME is the first comprehensive full-spectrum video evaluation suite designed for multimodal LLMs, covering 900 videos ($\approx$254 hours) and 2,700 manually aligned QA pairs across 6 domains and 30 fine-grained tasks. The benchmark integrates multimodal video inputs, including frames, audio, and subtitles, and spans a diverse range of durations from short clips to hour-long content. It is designed to evaluate a model's holistic video understanding, including temporal grounding, auditory reasoning, and long-range visual dependency modeling.

\paragraph{CVBench~\cite{zhu2025cvbench}.}
CVBench is a benchmark targeting cross-video relational reasoning, where multimodal models must jointly analyze multiple video streams to infer object associations, temporal relationships, and event-level dependencies. It contains approximately 1,000 carefully annotated QA pairs spanning object-level association, event matching, cross-video correspondence, and higher-order reasoning. The benchmark specifically evaluates a model’s ability to integrate visual evidence across disjoint videos, an ability that current MLLMs often lack due to their limited temporal relation modeling capability.

\paragraph{STI-Bench~\cite{li2025stibench}.}
STI-Bench focuses on fine-grained spatial-temporal intelligence in real-world scenarios such as robot manipulation or vehicle operation. It evaluates whether models can accurately reason about object appearance, pose, displacement, and dynamic state changes across desktop, indoor, and outdoor videos. The benchmark emphasizes physically grounded prediction, challenging models to infer where objects were, are, and will be and revealing substantial limitations in spatial-temporal consistency of current VLMs.

\paragraph{DSI-Bench~\cite{zhang2025dsibench}}
DSI-Bench is designed to systematically measure dynamic spatial intelligence by decomposing 3D spatial reasoning into independent motion factors. It contains nearly 1,000 dynamic videos and over 1,700 expert-verified QA pairs, covering nine fundamental motion patterns involving both moving observers and moving objects. By isolating each motion factor, such as ego-motion, object translation, rotation, and coordinated interaction, the benchmark provides a principled way to diagnose which dimensions of spatial reasoning current models fail to capture.

\paragraph{OpenEQA~\cite{majumdar2024openeqa}}
OpenEQA is a large-scale benchmark designed for embodied question answering (EQA) in the era of foundation models, aiming to measure how well agents can perceive, navigate, and reason within 3D embodied environments. 
It provides more than 1,000 expert-verified EQA samples across diverse scenes and question types, covering perception, spatial reasoning, object grounding, action planning, and commonsense reasoning.

\subsection{Baselines}
Besides, the general VLM baselines (e.g., LLaVA-v1.5~\cite{liu2024improved}, LLaVA-NeXT-Video~\cite{zhang2024llavanext-video}, LLaVA-Video~\cite{zhang2024videollava}, LLaVA-Onevision~\cite{li2024llavaonevision,an2025llavaonversion15}, InternVL~\cite {chen2025internvl, zhu2025internvl3}, Qwen2.5/3-VL~\cite{bai2025qwen25vl}), we also compared our methods with baselines that have been fine-tuned on the spatial-focused datasets, including
\begin{compactitem}
    \item \textbf{LLaVA3D}~\cite{zhu2025llava3d} is built upon the LLaVA-Video-7B~\cite{zhang2024videollava}, where the vision encoder is extended with a 3D-aware adapter that fuses RGB frames with 3D geometric cues. It is fine-tuned on large-scale 3D-annotated datasets such as LLaVA-3D-Instruct-86K~\cite{zhu2025llava3d}, and MMScan QA~\cite{lyu2024mmscan}.
    
    \item \textbf{SpaceR}~\cite{ouyang2025spacer} employs a Qwen-2.5-VL-7B-Instruct~\cite{bai2025qwen25vl} as the backbone. The model is fine-tuned on a carefully cured dataset, i.e., SpaceR-151k.
    
    \item \textbf{SpatialMLLM}~\cite{wu2025spatialmllm} adopts the Qwen2.5-VL-3B~\cite{bai2025qwen25vl} as backbone and VGGT~\cite{wang2025vggt} as the spatial encoder. It is trained on the Spatial-MLLM-120k.
    
    \item \textbf{Video3D-LLM}~\cite{zheng2025video3dllm} builds on the LLaVA-Video-7B~\cite{zhang2024videollava}. The model is fine-tuned on 3D-awared datasets, including ScanRefer~\cite{chen2020scanrefer}, Multi3DRefer~\cite{zhang2023multi3drefer}, Scan2Cap~\cite{chen2021scan2cap}, ScanQA~\cite{azuma2022scanqa} and SQA3D~\cite{ma2022sqa3d}.
    
    \item \textbf{Ross3D}~\cite{wang2025ross3d} is developed based on LLaVA-Video-7B~\cite{zhang2024videollava}, with cross-view and global-view reconstructions, enabling accurate spatial relationship modeling and comprehensive scene layout comprehension. It is fine-tuned on the combination of training sets of SQA3D~\cite{ma2022sqa3d}, ScanQA~\cite{azuma2022scanqa}, Scan2Cap~\cite{chen2021scan2cap}, ScanRefer~\cite{chen2020scanrefer}, and Multi3DRefer~\cite{zhang2023multi3drefer} datasets.
\end{compactitem}

\section{More Experimental Results}
\label{app:experimental_results}

\subsection{Detailed Benchmark Results}
We provide more detailed results on the existing benchmark in Table~\ref{tab:mmstar}, \ref{tab:3dsrnench}, \ref{tab:mmsibench}, \ref{tab:vsibench}, \ref{tab:cvbench}, \ref{tab:videomme}, and \ref{tab:openeqa}.
Moreover, we also conduct experiments on more focus on cross-vision synergy benchmark, including CV-Bench~\cite{tong2024cambrian}, STI-Bench~\cite{li2025stibench}, and DSI-Bench~\cite{zhang2025dsibench}. 
As results demonstrated in Table~\ref{tab:cv_bench}, \ref{tab:stibench}, and \ref{tab:dsibench}, the proposed PolyV consistently achieves superior average performance compared with all baselines, demonstrating its strong cross-vision synergistic reasoning capability.

\begin{table*}[!ht]
\centering
\caption{Detailed evaluation results on \textbf{MMSI-Bench}~\cite{yang2025mmsibench}. `Cam.' denotes the camera, `Obj.' means the object, `Reg.' is the region, `Meas.' is the measurement, and `Appr.' is the appearance. }
\label{tab:mmsibench}
\vspace{-2mm}
\fontsize{9}{11}\selectfont
\setlength{\tabcolsep}{0.5mm}
\begin{tabular}{lcccccccccccc}
\toprule
\multirow{2}{*}{Model} & \multicolumn{6}{c}{Positional Relationship} & \multicolumn{2}{c}{Attribute} & \multicolumn{2}{c}{Motion} & \multirow{2}{*}{MSR} & \multirow{2}{*}{Average} \\
\cmidrule(r){2-7}\cmidrule(r){8-9}\cmidrule(r){10-11}
& Cam.–Cam. & Obj.–Obj.  & Reg.–Reg.  & Cam.–Obj.  & Obj.–Reg.  & Cam.–Reg.  & Meas. &  Appr.  & Cam.  & Obj. & &  \\
\midrule
LLaVA-v1.5-7B~\cite{liu2024llava15} & 33.3 & 31.9 & 22.2 & 29.1 & 22.3 & 20.5 & 25.0 & 22.7 & 14.9 & 21.0 & 22.2 & 24.2 \\
LLaVA-NeXT-Video-7B~\cite{zhang2024llavanext-video} & 21.5 & 22.3 & 24.7 & 26.7 & 23.5 & 22.9 & 32.8 & 27.3 & 16.2 & 26.3 & 27.8 & 24.9\\
Qwen2.5-VL-7B~\cite{bai2025qwen25vl} &  24.7 &  24.5 &  24.7 &  25.6 &  29.4 &  26.5 &  25.0 &  18.2 &  20.3 &  \bf 39.5 &  25.8 &  25.9 \\
Qwen3-VL-8B~\cite{bai2025qwen25vl} & 31.2 & 27.7 & 28.4 & 24.4 & 28.2 & \bf 34.9 & \bf 35.9 & 15.1 & \bf 27.0 & 38.2 & 28.3 & 29.0\\
InternVL2.5-8B~\cite{chen2025internvl}  & 32.3 &  27.7 &  29.6 &  32.6 &  24.7 &  32.5 &  26.6 &  27.3 &  16.2 &  31.6 &  \bf 30.3 &  28.7 \\
InternVL3-8B~\cite{zhu2025internvl3}  & 25.8 &  31.9 &  \bf 37.0 &  25.6 &  \bf 35.3 &  28.9 &  23.4 &  24.2 &  16.2 &  32.9 &  14.6 &  25.7 \\
LLaVA-OneVision-7B~\cite{li2024llavaonevision}  & 20.4  & \bf 33.0  & 29.6  & 29.1  & 25.9  & 30.1  & 29.7  & 25.8  & 18.9  & 34.2  & 11.6  & 24.5 \\
LLaVA-OneVision1.5-8B~\cite{an2025llavaonversion15} & \bf34.4 & 26.6 & 33.3 & 33.7 & 34.1 & 25.3 & 25.0 & 28.8 & 22.9 & 30.3 & 29.3 & 29.6\\
\hdashline 
LLaVA3D~\cite{zhu2025llava3d} & 18.3 & 18.1 & 16.0 & 15.1 & 21.2 & 22.9 & 29.7 & 24.2 & 24.3 & 27.6 & 18.7 & 20.8 \\
Ross3D~\cite{wang2025ross3d} & 12.9 & 12.8 & 13.6 & 11.6 & 23.5 & 25.3 & 31.2 & 6.06 & 5.4 & 23.7 & 16.2 & 16.4 \\
SpaceR~\cite{ouyang2025spacer} & 29.0 & 26.6 & 24.7 & 30.2 & 30.6 & 27.7 & 23.4 & 21.2 & 20.3 & 34.2 & 27.8 & 27.2\\
SpatialMLLM~\cite{wu2025spatialmllm} & 25.8 & 25.5 & 28.4 & 34.9 & 24.7 & 25.3 & 18.7 & 33.3 & 12.2 & 23.7 & 26.8 & 25.7 \\
Video3D-LLM~\cite{zheng2025video3dllm} & 15.0 & 4.3 & 6.2 & 1.2 & 9.4 & 2.4 & 1.6 & 3.0 & 1.3 & 9.2 & 7.1 & 5.9 \\
\hdashline
\rowcolor{lightblue} PolyV (Ours) & 33.0 & 30.0 & 36.0 & 33.0 & 34.0 & 30.0 & 28.0 & \bf 30.0 & 26.0 & 38.0 & 30.0 &  \bf 31.7 \\
\bottomrule
\end{tabular}
\end{table*}

\begin{table*}[!ht]
\centering
\caption{Detailed evaluation results on \textbf{VSI-Bench}~\cite{yang2025vsibench}. object count (Obj. Count), absolute distance (Abs. Dist.), object size (Obj. Size), room size, relative distance (Rel. Dist.), relative direction (Rel. Dir.), route plan, appearance order (Appr. Order)}
\label{tab:vsibench}
\vspace{-2mm}
\fontsize{9}{11}\selectfont
\setlength{\tabcolsep}{1.0mm}
\begin{tabular}{lccccccccc}
\toprule
Model & Average & Obj. Count & Abs. Dist. &  Obj. Size &   Room Size &   Rel. Dist. &   Rel. Dir. &  Route Plan  &  Appr. Order\\
\midrule
LLaVA-Video-7B~\cite{zhang2024videollava} & 35.6 &  48.5 &  14.0  & 47.8 &  24.2 &  43.5 &  42.4 &  34.0 &  30.6 \\
LLaVA-NeXT-Video-7B~\cite{zhang2024llavanext-video} & 29.9 & 53.6 & 30.6  & 21.6 & 38.8  & 27.7 & 33.9 & 31.4  & 16.1 \\
LLaVA-OneVision-7B~\cite{li2024llavaonevision} & 32.4 &  47.7 &  20.2 &  47.4 &  12.3 &  42.5 &  35.2  & 29.4  & 24.4 \\
LLaVA-OneVision1.5-8B~\cite{an2025llavaonversion15}&  46.3  & 71.2 & 35.1 & \bf 69.3  & 60.6 & 38.3  & 39.1 & 29.9 & 25.2 \\
InternVL3-8B~\cite{zhu2025internvl3}  & 50.7 & \bf 75.8 & 52.2 &65.2  & \bf 70.4 & 36.8 & 36.4  & 37.3 & \bf 37.2 \\
\hdashline
Video3d-LLM~\cite{zheng2025video3dllm} & 15.8 & 17.5 & 0.0 & 0.0 & 0.0 & 29.6 & 33.4 & 26.3 & 24.8 \\
SpaceR~\cite{ouyang2025spacer} &24.5 & 33.1 & 0.0 & 2.5 & 0.0 & 41.7 & \bf 44.7 & 32.0 & 46.4  \\
LLaVA3D~\cite{zhu2025llava3d} & 9.4 & 2.5 & 0.0 & 0.1 & 0.0 & 31.8 & 6.1 & 27.8 & 18.9 \\
\hdashline
\rowcolor{lightblue} PolyV (Ours) & \bf 52.7  & 74.6 & \bf 56.0 & 65.0 & 70.0 & \bf 45.0 & 42.0 & \bf 40.0 & 29.0 \\
\bottomrule
\end{tabular}
\end{table*}

\begin{table*}[!ht]
\centering
\caption{Detailed evaluation results on \textbf{CVBench}~\cite{zhu2025cvbench}. The tasks include: multi-view scene understanding (M. SU), multi-video temporal reasoning (M. TR), joint-video spatial navigation (J. SN), video difference captioning (VDC), cross-video counterfactual reasoning (C. CR), joint-video summarization (J.S), and cross-video procedural transfer (C. PT). }
\label{tab:cvbench}
\vspace{-2mm}
\fontsize{9}{11}\selectfont
\setlength{\tabcolsep}{4.0mm}
\begin{tabular}{lcccccccc}
\toprule
Model& Average & M. SU & M. TR & J. SN & VDC  & C.CR  & J.S  & C.PT \\
\midrule
LLaVA-NeXT-Video-7B~\cite{zhang2024llavanext-video} & 30.9 & 5.3 & 14.6 & 14.6 & 20.0 & 17.3  & 5.8 & 7.8\\
 Qwen2.5-VL-7B~\cite{bai2025qwen25vl}    &   51.3 &  80.0 &  22.7 &  26.2 &  50.9 &  55.8 &   69.2 &  60.8 \\
 InternVL2.5-8B~\cite{chen2025internvl} & \bf 59.4 &  83.6 &  26.7 &  50.0 &  60.0 &   \bf 69.2 &  67.3 &  \bf 68.6 \\
 Qwen2.5-VL-8B~\cite{bai2025qwen25vl} & 51.1 & 80.0 & 31.1 & 47.6 & \bf 71.1 & 66.7 &  71.1 &  68.0 \\
 LLaVA-OneVision-7B~\cite{li2024llavaonevision}    &  52.6 &  83.6 &  \bf 40.0 &  38.1 &  45.5 &  42.3 &  61.5 &  52.9 \\
 LLaVA-OneVision1.5-8B~\cite{an2025llavaonversion15} & 43.3 & 7.5 & 23.4 & 23.4 & 28.8 &27.7 & 8.7 & 12.5 \\
\hdashline
Video3D-LLM~\cite{zheng2025video3dllm} & 44.6 & 61.5 & 32.9 & 25.0 & 44.0 & 42.9 & 56.9 & 70.0 \\
Spatial-MLLM~\cite{wu2025spatialmllm} & 38.2  & 74.5 & 29.7 & 33.3 & 41.8 & 46.0 & 57.7 & 46.0 \\
SpaceR~\cite{ouyang2025spacer} & 50.4 & 83.6 & 31.1 & 26.2 & 49.1 & 54.9 & \bf 76.9 & 58.0  \\
Ross3D~\cite{wang2025ross3d} & 42.1 & 73.7 & 30.4 & 35.9 & 41.2 & 62.5 &  72.4 & 48.0\\
\hdashline
\rowcolor{lightblue}PolyV (Ours)  &  59.1 & \bf 86.0 & 38.0 & \bf 52.0 & 58.0 & 63.0 & 68.0 & 49.0\\
\bottomrule
\end{tabular}
\end{table*}

\begin{table*}[!ht]
\centering
\caption{Detailed evaluation results on \textbf{OpenEQA}$_\text{HM3D}$~\cite{majumdar2024openeqa}. `Obj. Rec.' denotes object recognition, `Attr. Rec.' is attribute recognition, `Spatial Und.' is the spatial understanding, `Obj. State Rec.' is the object state recognition, `Func. Rea.' is the function reasoning, `World Know.' is the world knowledge, ` Obj. Loc.' is the object localization. }
\label{tab:openeqa}
\vspace{-2mm}
\fontsize{9}{11}\selectfont
\setlength{\tabcolsep}{1.2mm}
\begin{tabular}{lccccccccc}
\toprule

Model & Obj. Rec. &  Attr. Rec. &  Spatial Und. &  Obj. State Rec. &  Func. Rea. &  World Know. & Obj. Loc. &  Average \\
\midrule
LLaVA-NeXT-Video-7B~\cite{zhang2024llavanext-video} & 	31.5 & 34.1 & 35.0 & 31.2 & 51.0 & 31.7 & 30.2 & 34.9  \\
LLaVA-OneVision-7B~\cite{li2024llavaonevision} & 35.1 & 42.5 & 33.8 & 35.0 & 53.7 & 35.9 & 34.6 & 38.7 \\
LLaVA-OneVision1.5-8B~\cite{an2025llavaonversion15} &37.5 & 45.6 & 36.8 & 37.2 & 58.9 & 38.0 & 36.4 & 41.2 \\
InternVL2.5-8B~\cite{chen2025internvl} & 42.3 & 51.0 & 40.0 & 41.3 & 68.0 & 41.5 & 40.5 & 46.7 \\
InternVL3-8B~\cite{chen2025internvl} & 44.4 & 51.1 & 42.5 & 43.5 & \bf 68.7 & 43.6 & 42.5 & 48.0  \\
Qwen2.5-VL-7B~\cite{bai2025qwen25vl} &  50.4 & 62.1 & 49.3 & 66.4 & 61.4 & 60.3 & 47.6 & 56.6  \\
\hdashline
Spatial-MLLM~\cite{wu2025spatialmllm} &  33.7 & 9.7 & 15.5 & 1.7 & 31.2 & 21.2 & 17.0 & 18.0 \\
Video3d-LLM~\cite{zheng2025video3dllm} & 33.2 & 41.2 & 47.0 & 60.0 & 45.5 & 43.7 & 35.7 & 43.7 \\
Ross3D~\cite{wang2025ross3d} & 41.7 & 59.5 & 52.5 & 64.2 & 57.5 & 55.5 & 39.7 & 52.5 \\
\hdashline
\rowcolor{lightblue}PolyV (Ours)  &  \bf 60.8 & \bf 66.7 & \bf 57.9 & \bf 70.4 & 68.6 & 61.8 & \bf 58.2 &  \bf 63.4 \\
\bottomrule
\end{tabular}
\end{table*}

\begin{table}[!ht]
\centering
\caption{Detailed evaluation results on \textbf{VideoMME}$_\text{w/o sub}$~\cite{fu2025videomme}.}
\label{tab:videomme}
\vspace{-2mm}
\fontsize{9}{11}\selectfont
\setlength{\tabcolsep}{1.5mm}
\begin{tabular}{lcccc}
\toprule
Model   & Short & Mid & Long & Average\\
\midrule
LLaVA-NeXT-Video-34B~\cite{zhang2024llavanext-video}  & 61.7 &  50.1 & 44.3 &  52.0\\ 
Qwen3-VL-8B~\cite{bai2025qwen25vl} & 58.1 & 49.2 & 49.1 & 52.1 \\
\hdashline
LLaVA-3D~\cite{zhu2025llava3d} & 27.1 & 27.0 & 25.4 & 26.5 \\
SpaceR~\cite{ouyang2025spacer} & 67.0 & 55.4 & 48.2 & 56.9 \\
Spatial-MLLM~\cite{wu2025spatialmllm} & 54.4 & 42.0 & 35.8 & 44.1 \\
Video3d-LLM~\cite{zheng2025video3dllm} & 47.7 & 42.0 & 37.1 & 41.9 \\
Ross3D~\cite{wang2025ross3d} & 44.1 & 39.9 & 38.9 & 40.9 \\
\hdashline
\rowcolor{lightblue}PolyV (Ours)  &\bf 75.2 & \bf 70.1 & \bf 63.5 & \bf 69.6\\
\bottomrule
\end{tabular}
\end{table}

\begin{table}[!ht]
\centering
\caption{Detailed evaluation results on \textbf{CV-Bench}~\cite{tong2024cambrian}.}
\label{tab:cv_bench}
\vspace{-2mm}
\fontsize{9}{11}\selectfont
\setlength{\tabcolsep}{0.7mm}
\begin{tabular}{lccccc}
\toprule
\multirow{2}{*}{Model}   & \multicolumn{2}{c}{2D} & \multicolumn{2}{c}{3D} & \multirow{2}{*}{Average} \\
\cmidrule(r){2-3}\cmidrule(r){4-5}
& Relation & Count & Depth & Distance &  \\
\midrule
Qwen2.5-VL-7B~\cite{bai2025qwen25vl}  & 77.4 & 63.2 & 74.2 & 44.0 & 64.8 \\ 
Qwen3-VL-8B~\cite{bai2025qwen25vl} & 92.5 & 71.7 & 95.0 & 58.7 & 79.1  \\
\hdashline
SpaceR~\cite{ouyang2025spacer} & 83.7 & 65.1 & 78.5 & \bf 66.0 & 72.9 \\
Spatial-MLLM~\cite{wu2025spatialmllm} & 62.3 & 53.7 & 60.5 & 65.7 & 60.1\\
Video3d-LLM~\cite{zheng2025video3dllm} & 53.2 & 18.4 & 56.2 & 60.5 & 45.1  \\
Ross3D~\cite{wang2025ross3d} & 13.8 & 6.3 & 33.2 & 24.2 & 18.3 \\
\hdashline
\rowcolor{lightblue}PolyV (Ours)  & \bf 96.0 &\bf  82.0 & \bf 98.0 & 58.0 & \bf  83.5 \\
\bottomrule
\end{tabular}
\end{table}

\begin{table*}[!ht]
\centering
\caption{Detailed evaluation results on \textbf{DSI-Bench}~\cite{zhang2025dsibench} across different video sources.}
\label{tab:dsibench}
\vspace{-2mm}
\fontsize{9}{11}\selectfont
\setlength{\tabcolsep}{3.5mm}
\begin{tabular}{lcccccc}
\toprule
{Model}   & CamerBench~\cite{lin2025towards} &	SynFMC~\cite{shuai2025free} &		internet &		k700~\cite{smaira2020short} &		llava178k~\cite{zhang2024videollava} &		Average \\
\midrule
LLaVA-NeXT-Video-7B~\cite{zhang2024llavanext-video} & 	36.0 & 	28.4 & 	36.9 & 	33.6 & 	35.8 & 	35.6  \\
LLaVA-OneVision-7B~\cite{li2024llavaonevision} & 	46.0 & 	33.0 & 	48.1 & 	47.5 & 	45.3 & 	46.4 \\
LLaVA-OneVision1.5-8B~\cite{an2025llavaonversion15} &	50.4 & 	48.6 & 	50.3 & 	57.0 & 	\bf 49.6 & 	51.1 \\
InternVL2.5-8B~\cite{chen2025internvl} & 	47.6 & 	48.6 & 	50.4 & 	57.4 & 	46.8 & 	50.4 \\
InternVL3-8B~\cite{chen2025internvl} & 	52.8 & 	43.1 & 	45.9 & 	47.5 & 	47.6 & 	47.2 \\
Qwen2.5-VL-7B~\cite{bai2025qwen25vl}  & 40.0 & 38.5 & 39.4 & 41.2 & 48.1 & 40.5  \\ 
Qwen3-VL-8B~\cite{bai2025qwen25vl} & 38.3 & 37.6 & 39.5 & 37.4 & 40.3 & 39.0  \\
\hdashline
SpaceR~\cite{ouyang2025spacer} & 39.4 & 41.3 & 39.5 & 43.1 & 43.6 & 40.7 \\
Spatial-MLLM~\cite{wu2025spatialmllm} & 27.5 & 24.8 & 30.7 & 29.8 & 31.2 & 29.8\\
Video3d-LLM~\cite{zheng2025video3dllm} & 39.9 & 34.8 & 38.9 & 49.6 & 38.2 & 40.0\\
Ross3D~\cite{wang2025ross3d} & 37.1 & 32.3 & 34.3 & 37.3 & 30.8 & 34.5 \\
\hdashline
\rowcolor{lightblue}PolyV (Ours)  & \bf  60.8 & \bf 57.4 & \bf 59.3 & \bf 67.9 & 48.1 & \bf 58.7\\
\bottomrule
\end{tabular}
\end{table*}

\begin{table*}[!ht]
\centering
\caption{Detailed evaluation results on \textbf{STI-Bench}~\cite{li2025stibench}. `Dim. Meas' is the Dimensional Measurement, `Disp. \& P.L.' means Displacement \& Path Length, `Speed \& Acc.' denotes Speed \& Acceleration, `Ego Orient.' indicates Ego-Centric Orientation, `Traj. Desc.' is Trajectory Description, and `Pose Est.' is Pose Estimation.}
\label{tab:stibench}
\vspace{-2mm}
\fontsize{8}{11}\selectfont
\setlength{\tabcolsep}{0.5mm}
\begin{tabular}{lccccccccc}
\toprule
\multirow{2}{*}{Model}   & \multicolumn{3}{c}{Static Understanding} &	\multicolumn{5}{c}{Dynamic Understanding}&		\multirow{2}{*}{Average} \\
\cmidrule(r){2-4}\cmidrule(r){5-9}
& Dim. Meas &  Spatial Relation &  3D Video Grounding &  Disp. \& P.L. &  Speed \& Acc. &  Ego Orient.  & Traj. Desc. &  Pose Est. \\
\midrule
LLaVA-NeXT-Video-7B~\cite{zhang2024llavanext-video} & 	22.1 & 42.5 & 19.9 & 21.5 & 21.1 & 27.0 & 42.3 & 18.9 & 23.6 \\
LLaVA-OneVision-7B~\cite{li2024llavaonevision} & 	25.6 & 30.8 & 24.6 & 25.1 & 30.8 & 41.6 & 51.3 & \bf 55.8 & 34.2\\
LLaVA-OneVision1.5-8B~\cite{an2025llavaonversion15} &\bf 34.3 & 29.4 & 33.1 & 24.9 & \bf 35.0 & 63.8 & 44.9 & 55.6 & 39.0   \\
InternVL2.5-8B~\cite{chen2025internvl} & 	26.3 & 50.7 & 31.5 & 20.9 & 29.3 & \bf 66.5 & 46.1 & 56.7 & 38.0\\
InternVL3-8B~\cite{chen2025internvl} & 26.3 & 52.0 & 27.1 & 21.5 & 33.2 & 0.2 & 25.6 & 40.6 & 30.4 \\
Qwen2.5-VL-7B~\cite{bai2025qwen25vl}  & 24.5 & 46.6 & 32.8 & 24.6 & 30.6 & 17.8 & 43.6 & 47.2 & 32.4 \\ 
Qwen3-VL-8B~\cite{bai2025qwen25vl} & 25.9 & 50.7 & 37.8 & 19.6 & 26.7 & 55.7 & 39.7 & 56.1 & 37.0 \\
\hdashline
SpaceR~\cite{ouyang2025spacer} & 21.4 & 47.3 & 30.3 & 23.0 & 29.4 & 26.5 & 47.4 & 49.2 & 32.4 \\
Spatial-MLLM~\cite{wu2025spatialmllm} & 23.5 & 42.5 & 17.0 & 17.9 & 12.1 & 51.3 & 28.2 & 26.9 & 24.3 \\
Video3d-LLM~\cite{zheng2025video3dllm} &22.5 & 42.5 & 20.5 & 19.3 & 23.9 & 13.5 & 39.7 & 17.2 & 22.2 \\
\hdashline
\rowcolor{lightblue}PolyV (Ours)  & 33.4 &\bf  55.8 & \bf 40.1 & \bf 26.3 & 34.7 & 59.2 & \bf 48.6 & 36.3 & \bf 46.8\\
\bottomrule
\end{tabular}
\end{table*}

\subsection{Mitigating Catastrophic Forgetting Across Modalities}
Our experiments reveal that models fine-tuned solely on spatial-centric data tend to suffer from catastrophic forgetting of previously acquired visual knowledge.
For example, although SpaceR and Qwen2.5-VL-7B share the same backbone, SpaceR exhibits noticeably degraded performance on most benchmarks after training, indicating that its incremental learning strategy inadvertently erodes the image- and video-based competencies learned during pre-training.
Similar degradation is observed for LLaVA-Video and other 3D-aware models when evaluated on VSI-Bench (cf. Table~\cite{yang2025vsibench}), suggesting that current approaches struggle to balance the acquisition of new spatial capabilities with the retention of prior modality knowledge.
In contrast, our proposed PolyV effectively avoids such forgetting. After learning from diverse modality-specific data, PolyV not only maintains strong performance on original image- and video-based benchmarks but also successfully incorporates additional spatial understanding. This demonstrates its ability to achieve truly synergistic multi-modality learning without sacrificing previously learned capabilities.

\begin{figure*}[!ht]
    \centering
    \includegraphics[width=0.99\linewidth]{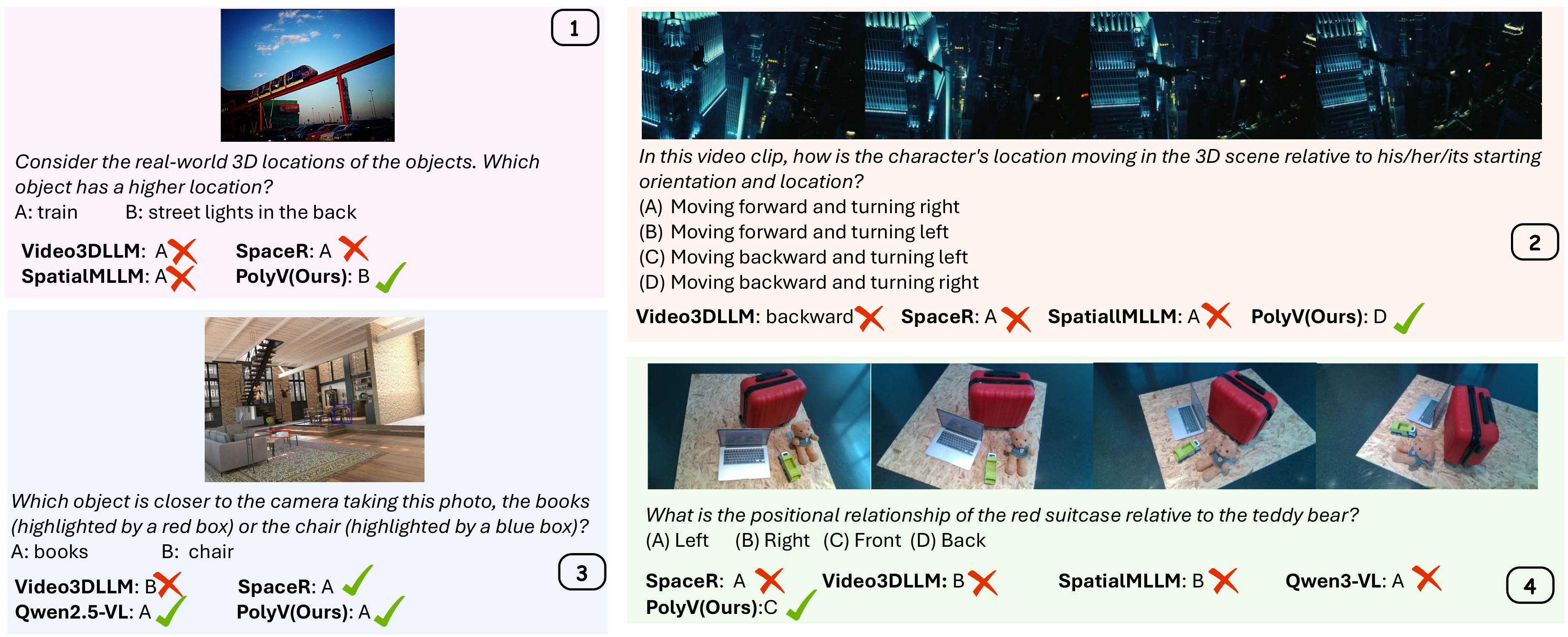}
    \caption{Qualitative comparisons between PolyV and existing models. Case 1 is from 3DSRBench~\cite{ma20253dsrbench}, Case 2 from DSI-Bench~\cite{zhang2025dsibench}, Case 3 from CV-Bench~\cite{tong2024cambrian}, and Case 4 from STI-Bench~\cite{li2025stibench}.}
    \label{fig:case2}
\end{figure*}

\subsection{More Quantitation Results}
We provide more quantitation results in Fig.~\ref{fig:case2}.

\section{Limitation and Future Work}
Although we believe that synesthetic, cross-vision synergy is a fundamental capability that future vision–language models must possess—especially for downstream applications such as robotics and autonomous driving—our current approach still faces several limitations.
First, despite the use of a MoE architecture, which reduces computational cost during deployment by activating only a subset of experts, the overall model remains resource-intensive. Efficient scaling and lightweight deployment thus remain open challenges.
Second, video and 3D processing inherently require handling large numbers of frames or point-cloud tokens. Current input-length constraints limit the model’s ability to fully capture long-term dynamics or fine-grained spatial structures.
Furthermore, our model acquires synesthetic ability primarily through supervised fine-tuning. While effective, this approach may not fully exploit the model’s potential for self-thinking.

For future work, we plan to explore reinforcement learning–based cross-vision synergy. By encouraging the model to explicitly reason about spatial structure during its thinking process, e.g., inferring 3D geometry from 2D images or predicting spatial evolution from video cues, the model may develop more robust, self-refining synesthetic abilities. Such advances are expected to yield stronger performance across a wide spectrum of vision-centric downstream tasks.

\end{document}